\documentclass[10pt,twocolumn,letterpaper]{article}

\usepackage{iccv}
\usepackage{times}
\usepackage{epsfig}
\usepackage{graphicx}
\usepackage{amsmath}
\usepackage{amssymb}

\usepackage{marvosym}
\usepackage{booktabs}
\usepackage[linesnumbered,ruled,vlined]{algorithm2e}
\usepackage{colortbl}
\usepackage[accsupp]{axessibility}

\usepackage[pagebackref=true,breaklinks=true,letterpaper=true,colorlinks,bookmarks=false]{hyperref}

\usepackage[capitalize]{cleveref}
\crefname{section}{Sec.}{Secs.}
\Crefname{section}{Section}{Sections}
\Crefname{table}{Table}{Tables}
\crefname{table}{Tab.}{Tabs.}

\iccvfinalcopy

\def\iccvPaperID{8973} % *** Enter the ICCV Paper ID here

\ificcvfinal\fi

\begin{document}

\title{Global Adaptation meets Local Generalization: Unsupervised Domain Adaptation for 3D Human Pose Estimation}

\author{
Wenhao Chai\textsuperscript{1} \quad
Zhongyu Jiang\textsuperscript{2} \quad
Jenq-Neng Hwang\textsuperscript{2} \quad
Gaoang Wang\textsuperscript{1 \Letter}\\
[2mm]
$^1$~Zhejiang University \quad $^2$~University of Washington\\
% [2mm]
% {\tt\small \{wenhaochai.19, gaoangwang\}@intl.zju.edu.cn, \{zyjiang, hwang\}@uw.edu}
}

\maketitle

\begin{abstract}

When applying a pre-trained 2D-to-3D human pose lifting model to a target unseen dataset, large performance degradation is commonly encountered due to domain shift issues.
We observe that the degradation is caused by two factors: 1) the large distribution gap over global positions of poses between the source and target datasets due to variant camera parameters and settings, and 2) the deficient diversity of local structures of poses in training.
To this end, we combine \textbf{global adaptation} and \textbf{local generalization} in \textit{PoseDA}, a simple yet effective framework of unsupervised domain adaptation for 3D human pose estimation.
Specifically, global adaptation aims to align global positions of poses from the source domain to the target domain with a proposed global position alignment (GPA) module. And local generalization is designed to enhance the diversity of 2D-3D pose mapping with a local pose augmentation (LPA) module. These modules bring significant performance improvement without introducing additional learnable parameters. In addition, we propose local pose augmentation (LPA) to enhance the diversity of 3D poses following an adversarial training scheme consisting of 1) a augmentation generator that generates the parameters of pre-defined pose transformations and 2) an anchor discriminator to ensure the reality and quality of the augmented data. 
Our approach can be applicable to almost all 2D-3D lifting models. \textit{PoseDA} achieves 61.3 mm of MPJPE on MPI-INF-3DHP under a cross-dataset evaluation setup, improving upon the previous state-of-the-art method by 10.2\%.

\end{abstract}

\section{Introduction}
\label{sec:intro}

\begin{figure}
    \centering
    \includegraphics[width=.99\linewidth]{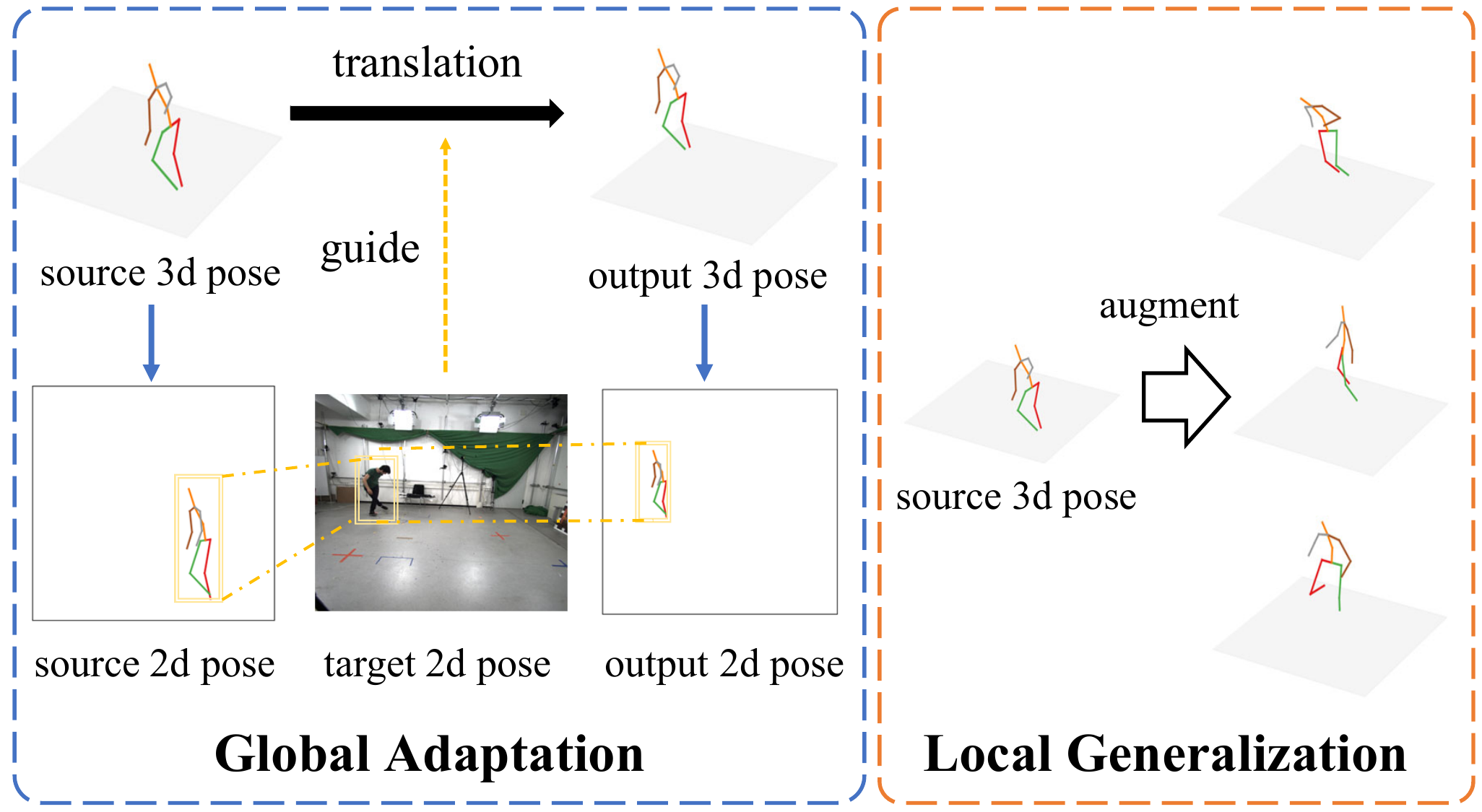}
    \caption{\small \textit{PoseDA} addresses 3D human pose domain adaptation problem through global adaptation and local generalization. The 2D poses from target dataset are used to guide the translation of 3D poses from source dataset. And the local root-relative poses also are augmented to achieve better generalization ability.}
    \label{fig:intro}
\end{figure}

% paragraph 1: task introduction
3D human pose estimation is an essential computer vision task which aims to estimate the coordinates of 3D joints from single-frame images or videos. This task can be further used for several downstream tasks in multiple object tracking~\cite{andriluka2018posetrack,fan2022deep,fan2022human,hao2023divotrack}, person re-identification~\cite{su2017pose}, action recognition~\cite{song2021human}, robot~\cite{svenstrup2009pose}, human body reconstruction~\cite{fan2022furpe}, sports application~\cite{zhao2023survey}, \textit{etc.}. However, large-scale 3D-annotated datasets are hard to obtain. Existing methods are usually built on an off-the-shelf 2D pose estimators~\cite{chen2018cascaded, sun2019deep} following two-stage schemes.

% paragraph 2: challenge
Deep learning methods~\cite{martinez2017simple, pavllo20193d} have achieved great success in the mapping from 2D to 3D in the past decade. Despite their success in in-distribution data, these fully-supervised methods show poor performance in cross-dataset inference~\cite{gong2021poseaug}. We argue that the real bottleneck lies in the domain gap of 3D pose data rather than the 2D-3D lifting network architecture or training strategy. Existing datasets either lack enough diversity in laboratorial environments~\cite{ionescu2013human3} or lack adequate quantity and accuracy in the wild~\cite{mehta2017monocular} due to the complex visual condition~\cite{ye2021perceiving,ye2022towards,ye2022underwater,liu2022nighttime,liu2023nighthazeformer,jiang2023five}.
We model the pose domain gap in terms of global position and local pose separately shown in Figure~\ref{fig:intro}. As for the global position, the camera intrinsic and extrinsic parameters are completely different in different datasets, resulting in performance degradation in cross-dataset evaluation. And for the local pose, the lack of action diversity also limits the generalization ability of the model.
Addressing the domain adaptation or generalization problem is a crucial step for 3D human pose estimation to move from toy experiments to real-world applications. Recent works focus more on enhancing the generalization ability of the 2D-3D lifting networks or set camera view prediction as an auxiliary task~\cite{wandt2019repnet, wang2020predicting} to address adaptation problems. Some methods apply data augmentation in training images through image transformations~\cite{rogez2016mocap, mehta2018single, mehta2017vnect} or human synthetics~\cite{chen2016synthesizing, hoffmann2019learning, varol2017learning}. However, our proposed method does not rely on RGB or temporal information. Specifically, our method first generates transformed pose pairs from source dataset and then use them to train the 2D-3D lifting network, thus can fit any off-the-shelf model.

% paragraph 4: our work
In this paper, we propose \textit{PoseDA}, an unsupervised domain adaptation framework for 3D human pose estimation. Our method only requires non-sequence 2D poses (not images) and camera intrinsic parameters in target dataset as well as a large-scale 3D-annotated human pose dataset (\textit{e.g.,}~Human3.6M~\cite{ionescu2013human3}). In real-world scenarios, obtaining prior knowledge of the camera intrinsic parameters is often not a concern. This is due to the fact that such information is readily available from camera specifications or can be inferred from the input images or videos alone~\cite{workman2015deepfocal}. The basic idea behind the proposed method is to combine global adaptation and local generalization to address the issue of unsupervised domain adaptation for 3D human pose estimation. Global adaptation aims to align global positions of poses from source domain to target domain, and local generalization aims to enhance the diversity of local structures of poses.
Therefore, our proposed method applies global position alignment strictly but only enhances the diversity of local poses. To be specific, we take a sample from the source dataset and apply transformations in terms of bone angle, bone length, and rotation. We use an augmentation generator to generate the parameters for these transformations and an anchor discriminator to ensure the realisticity and quality of these transformed pose pairs. As for the global position, we apply 2D global position alignment to ensure the alignment in both scales and 2D root positions between the projected 3D poses from the source domain with sampled 2D poses from the target domain. This process is solvable through geometry constraints with no additional learnable parameters. Finally, we use the transformed pose pairs to fine-tune the pre-trained 2D-3D lifting network and thus boost model performance on the target dataset without any use of 3D annotations.
Our contributions are summarized as follows:
\begin{itemize}
    \vspace{-5pt}
    \item We reduce the domain gap by separately applying global position alignment and local pose augmentation, where two major domain gaps are effectively decoupled.
    \vspace{-5pt}
    \item We align the global position through geometry constraint with no additional learnable parameters, which can boost model performance significantly. And we apply local pose augmentation to enhance the diversity of local structures of 3D poses.
    \vspace{-5pt}
    \item Our approach is applicable to almost all 2D-3D lifting models. We achieve the state-of-the-art performance in Human3.6M-3DHP cross-dataset evaluation with 61.3 mm of MPJPE.
\end{itemize}
\section{Related Work}
\label{sec:rel}
\subsection{Two-stage 3D Human Pose Estimation}
Inspired by the rapid development of 2D human pose estimation algorithms, many works have tried to utilize 2D pose estimation results for 3D human pose estimation to improve in-the-wild performance~\cite{wang2021deep, zhang2023mpm}. Two-stage 3D human pose estimation approaches, which first estimate 2D poses and then lift 2D poses to 3D poses, have been developed. Chen~\etal~\cite{chen20173d} present a simple approach to 3D human pose estimation by performing 2D pose estimation, followed by 3D exemplar matching. Martinez~\etal~\cite{martinez2017simple} propose a baseline focusing on lifting 2D poses to 3D with a simple yet effective neural network, which popularizes the research on lifting 2D pose to 3D space. Recently, semi/self-supervised learning based on geometry constraint~\cite{pavllo20193d, chen2019unsupervised, drover2018can, jiang2023back} has been used to train models without 3D annotations or by auxiliary losses.

\subsection{Data Augmentation on 3D Human Poses}
Data augmentation is widely used to improve deep model generalization ability by enhancing training data diversity. Some methods apply pose data augmentation on images~\cite{rogez2016mocap, mehta2017vnect} or generate 3D synthetic data using graphics engines~\cite{chen2016synthesizing, yang20183d, gong2022posetriplet}. Other methods directly generate 2D-3D pose pairs by applying transformations on 3D skeleton\cite{li2020cascaded, gong2021poseaug, gholami2022adaptpose}. Gong~\etal~\cite{gong2021poseaug} make this augmentation process differentiable and further learnable. Those transformations consist of bone angle, bone length, and rigid-body transformation. %We argue that rigid transformation is the most simple yet efficient.

\begin{figure*}[t]
    \centering
    \includegraphics[width=0.85\linewidth]{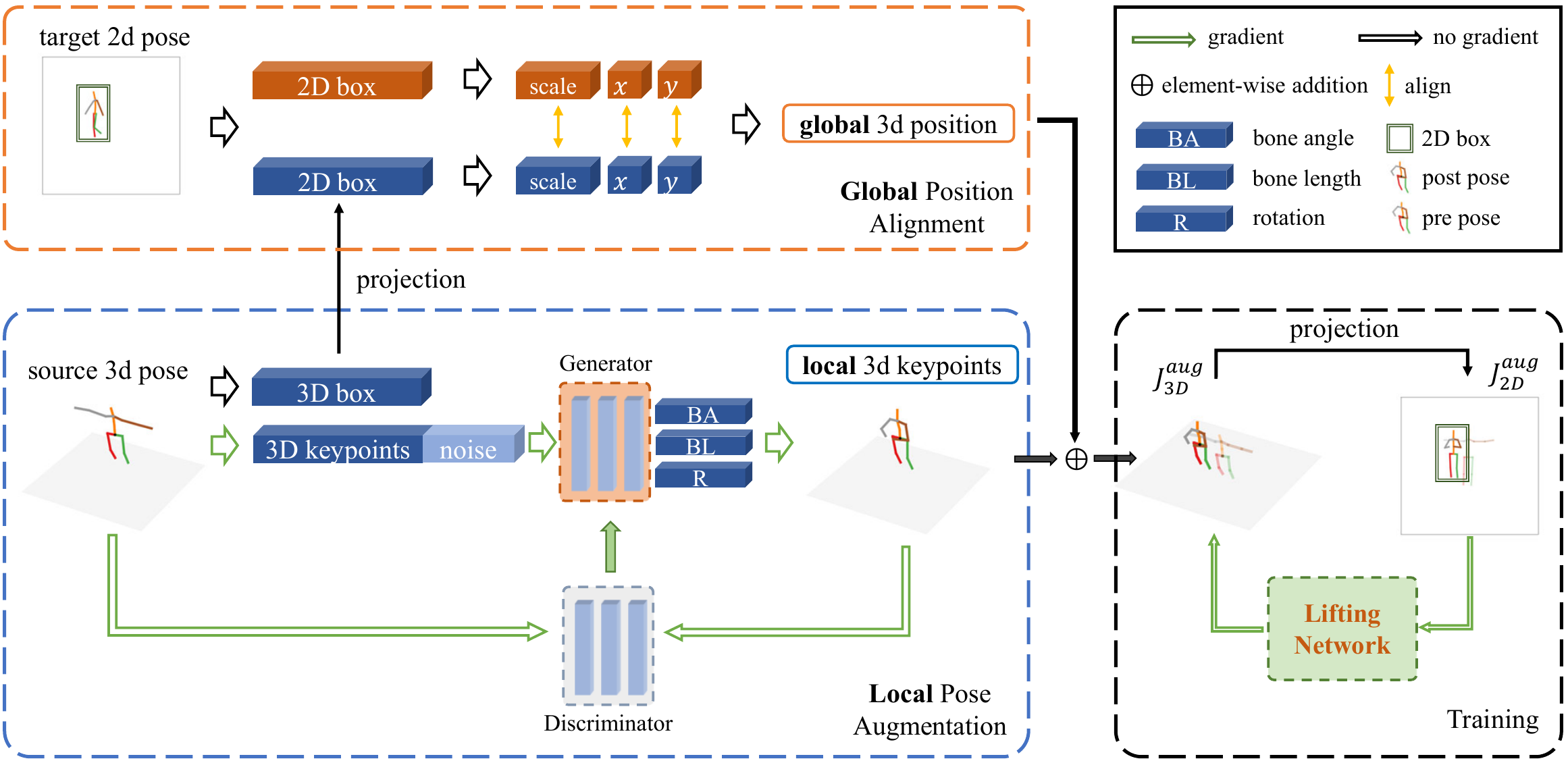}
    \caption{\small Our proposed unsupervised domain adaptation framework \textit{PoseDA} consists of global position alignment (GPA) and local pose augmentation (LPA). The augmentation bone angle (BA), bone length (BL), and rotation (R) are applied on source 3D poses through an adversarial augmentation framework consists of an augmentation generator $\mathcal{G}$ and an anchor discriminator $\mathcal{D}_{3D}$. Meanwhile, we also sample a 2D pose from target dataset and align the scale, x coordination, and y coordination in 2D screen between the target 2D box and projected source 2D box. Thus we solve the global 3D position by those geometric constraint. Finally, the augmented pose pairs combining with global 3D position and local 3D keypoints are used to train the lifting network $\mathcal{P}$.}
    \label{fig:overview}
\end{figure*}

\subsection{Unsupervised Domain Adaptation for 3D Human Pose Estimation} Unsupervised domain adaptation~\cite{wilson2020survey} aims to transfer models from a fully-labeled source domain to an unlabeled target domain. Kundu~\etal~\cite{kundu2022uncertainty} address unsupervised domain adaptation problem by modeling pose uncertainty based on RGB images. Li~\etal~\cite{li2020cascaded} are the first to generate corresponding 2D-3D pose pairs by applying skeleton transformations. Gong~\etal~\cite{gong2021poseaug} develop this data augmentation module with the differentiable form to jointly optimize the augmentation process with a end-to-end trained model. Gholami~\etal~\cite{gholami2022adaptpose} further utilize 2D pose in target domain to address the domain adaptation problem. 
However, all existing methods focus more on local pose augmentation or adaptation. In this paper, \textit{PoseDA} utilizes both local pose augmentation and global position alignment, and achieves state-of-the-art performance in cross-domain tasks.

%Many deep domain adaptation methods incorporate adversarial training. The Generative Adversarial Network (GAN) method~\cite{goodfellow2020generative} is a deep generative model that pits two networks generator and discriminator against one another. To close the domain gap, the discriminator is considered as a domain classifier to guide the generator to learn domain-invariant features~\cite{tzeng2017adversarial, long2018conditional}. This paradigm has proven to be fruitful in semantic segmentation~\cite{zou2018unsupervised, vu2019advent}, human pose estimation~\cite{yang20183d}, image de-raining~\cite{zhang2019image}, \textit{etc}.
\section{Method}
\label{sec:method}

\begin{figure*}
    \centering
    \includegraphics[width=0.8\linewidth]{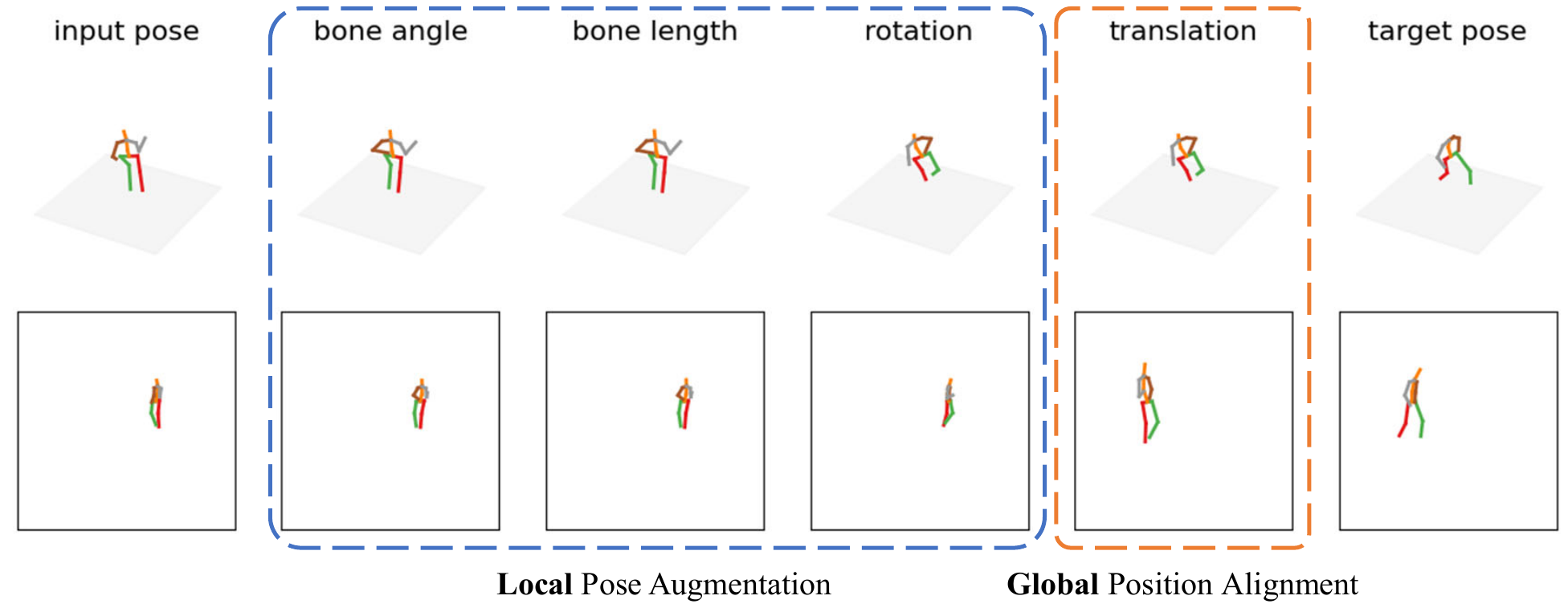}
    \caption{\small The pose transformations applied on 3D pose from source dataset~(column 1). Bone angle~(column 2), bone length~(column 3), and rotation~(column 4) are root-relative transformations given by local pose augmentation (LPA). Translation~(column 5) is given by global position alignment~(GPA). Target~(column 6) is the sampled 2D pose from target dataset used for GPA.}
    \label{fig:pipeline}
\end{figure*}

In this section, we introduce \textit{PoseDA}, an unsupervised domain adaptation framework, as summarized in Figure~\ref{fig:overview}, which consists of global position alignment (GPA) in Section~\ref{subsec:gpr} and local pose augmentation (LPA) in Section~\ref{subsec:ra}. Global position alignment module aims to align the 2D pose spatial distribution of both scale and location, $(x,y)$ coordinatse, between source and target datasets, and local pose augmentation is designed to enhance the diversity of 3D-2D pose mapping. Finally, we formulate the training process with several loss functions. The pseudo-code for the overall training process of our method is given in Algorithm~\ref{training pipeline}, and the transformation pipeline visualization of corresponding 2D-3D pose pairs is shown in Figure~\ref{fig:pipeline}.

\subsection{Problem Formulation}
\label{subsec:formulation}
% basic notation
Let $S=(\boldsymbol{J}^{src}_{2D}, \boldsymbol{J}^{src}_{3D})$ denote corresponding 2D-3D pose pairs from the source dataset, and $\boldsymbol{J}^{tar}_{2D}$ denote 2D pose from target dataset extracted by an off-the-shelf 2D pose estimator. Note that we only use $\boldsymbol{J}^{tar}_{3D}$ for evaluation but not for training. All the 3D poses  $\boldsymbol{J}_{3D}$ in this paper are root-relative since we do not predict the global position.

Our method conducts data augmentation on pose pairs $S=(\boldsymbol{J}^{src}_{2D}, \boldsymbol{J}^{src}_{3D})$ from source dataset. The 2D pose $\boldsymbol{J}^{tar}_{2D}$ and the camera intrinsic parameters from target dataset is used to guide this process:
\begin{equation}
    \boldsymbol{J}^{aug}_{3D} = \mathcal{G}(\boldsymbol{J}^{src}_{3D};\boldsymbol{\theta}_{cam}),\ 
    \boldsymbol{J}^{aug}_{2D} = f_p(\boldsymbol{J}^{aug}_{3D};\boldsymbol{\theta}_{cam})
\end{equation}
where $(\boldsymbol{J}^{aug}_{2D}, \boldsymbol{J}^{aug}_{3D})$ denotes augmented pose pairs, $\boldsymbol{\theta}_{cam}$ denotes the camera intrinsic parameters from target dataset, and $f_p$ denotes the projection function from 3D camera coordinates to 2D image coordinates. The augmented pose pairs are further used to train the pose lifting network.

We use several strong baseline methods~(\textit{e.g.,}~VideoPose3D~\cite{pavllo20193d} and SimpleBaseline~\cite{martinez2017simple}) for lifting 2D poses to 3D poses. Let $\mathcal{P}:\boldsymbol{J}_{2D} \mapsto \boldsymbol{J}_{3D}$ denotes the lifting network with parameters $\boldsymbol{\theta}_{\mathcal{P}}$, which can be trained in fully-supervised paradigm as:
\begin{equation}
    \min_{\boldsymbol{\theta}_{\mathcal{P}}} \mathcal{L}_{\mathcal{P}}(\mathcal{P}(\boldsymbol{J}_{2D};\boldsymbol{\theta}_{\mathcal{P}}), \boldsymbol{J}_{3D})
\end{equation}
where $(\boldsymbol{J}_{2D}, \boldsymbol{J}_{3D})$ denotes paired 2D-3D pose pairs, which consist of both augmented pose pairs $(\boldsymbol{J}^{aug}_{2D}, \boldsymbol{J}^{aug}_{3D})$ and original pose pairs $(\boldsymbol{J}^{src}_{2D}, \boldsymbol{J}^{src}_{3D})$, the loss function $\mathcal{L}_{\mathcal{P}}$ is typically defined as Mean Square Error (MSE), which is corresponding to the evaluation metric Mean Per Joint Position Error (MPJPE).

\subsection{Global Position Alignment (GPA)}
\label{subsec:gpr}
Global position alignment (GPA) is designed to eliminate the domain gap in viewpoints, which is simple yet efficient. By applying Monte Carlo sampling~\cite{chen2022epro}, the scale and location distributions of the 2D poses of the source dataset can be migrated to distributions of target dataset. We first construct pose pairs $(\boldsymbol{J}^{src}_{3D}, \boldsymbol{J}^{tar}_{2D})$, which are pairs of root-relative 3D poses and 2D poses randomly sampled from source and target domains respectively.

Given a 2D pose $\boldsymbol{J}^{tar}_{2D} = [\boldsymbol{x}^{tar}, \boldsymbol{y}^{tar}]^T \in \mathbb{R}^{2 \times J}$ and 3D pose $\boldsymbol{J}^{src}_{3D} = [\boldsymbol{X}^{src},\boldsymbol{Y}^{src},\boldsymbol{Z}^{src}]^T \in \mathbb{R}^{3 \times J}$ with the root at the origin $[0,0,0]^T$, GPA aims to estimate the translated 3D root position $\boldsymbol{J}_{r}= [ X_r , Y_r , Z_r]^T \in \mathbb{R}^{3 \times 1}$ to ensure the re-projected 2D pose $\boldsymbol{J}^{proj}_{2D} = [\boldsymbol{x}^{proj}, \boldsymbol{y}^{proj}]^T$ from $\boldsymbol{J}^{src}_{3D}+\boldsymbol{J}_r$ is close to $\boldsymbol{J}^{tar}_{2D}$ as much as possible with respect to both position and scale. 
Denote the operation of GPA as $\mathcal{F}$,
\begin{equation}
    \label{equ:GPR}
    \hat{\boldsymbol{J}}_{r} = \mathcal{F}(\boldsymbol{J}^{tar}_{2D}, \boldsymbol{J}^{src}_{3D}),
\end{equation}
where $\hat{\boldsymbol{J}}_{r}$ is the estimated 3D root position after translation.

We demonstrate how to obtain $\hat{\boldsymbol{J}}_{r}$ as follows. Assume that the camera intrinsic parameters are given. The projection from 3D joints to 2D joints after translation should obey the perspective projection function as follows:
\begin{equation}
\begin{aligned}
    \label{equ:p}
    & x_i^{proj} = \frac{f_x(X_i^{src}+X_r)}{Z_i^{src}+Z_r}+c_x, \\
    & y_i^{proj} = \frac{f_y(Y_i^{src}+Y_r)}{Z_i^{src}+Z_r}+c_y,
\end{aligned}
\end{equation}
% We assume that the camera intrinsics parameters are given, and the projection from 3D joints to 2D joints should obey the perspective projection function as follows:
% \begin{equation}
%     \label{equ:p}
%     x_i^{\prime}=\frac{x_i-c_x}{f_x} = \frac{X_i+X_r}{Z_i+Z_r}, y_i^{\prime}=\frac{y_i-c_y}{f_y} = \frac{Y_i+Y_r}{Z_i+Z_r}
% \end{equation}
where $i$ denotes the $i$-th joint, $(f_x, f_y), (c_x, c_y)$ denote focal length and principal point respectively. Note that $Z_i^{src}\ll Z_r$ since the absolute depth of the root joint of the person $Z_r$ is usually much larger than depth offset $Z_i^{src}$ of a certain joint relative to the root joint. Therefore, we assume that $Z_i^{src}+ Z_r \approx Z_r$ holds. Then Eq.~(\ref{equ:p}) becomes
\begin{equation}
\begin{aligned}
    \label{equ:p2}
    & x_i^{proj} \approx \frac{f_x(X_i^{src}+X_r)}{Z_r}+c_x, \\
    & y_i^{proj} \approx \frac{f_y(Y_i^{src}+Y_r)}{Z_r}+c_y,
\end{aligned}
\end{equation}

% Then we need to align $\boldsymbol{J}^{proj}_{2D}$ with $\boldsymbol{J}^{tar}_{2D}$ with respect to both scale and position. 
To achieve the similar scale with $\boldsymbol{J}^{tar}_{2D}$ for $\boldsymbol{J}^{proj}_{2D}$, we ensure that the 2D boxes should have similar size with the following constraint,
\begin{equation}
\label{equ:size}
\Delta x^{proj}+\Delta y^{proj} = \Delta x^{tar}+\Delta y^{tar},
\end{equation}
where $\Delta$ denotes the difference between the max-min coordinates of 2D joints, \textit{i.e.}, the width and height of the 2D box. Combine Eq.~{\ref{equ:p2}} and Eq.~{\ref{equ:size}}, we can get the approximated $Z_r$,
\begin{equation}
    \hat{Z}_r \approx \frac{f_x\Delta X^{src}+ f_y\Delta Y^{src}}{\Delta x^{tar} + \Delta y^{tar}}.
\end{equation}

Then, we can align the global (root) position between $\boldsymbol{J}^{proj}_{2D}$ and $\boldsymbol{J}^{tar}_{2D}$ by
\begin{equation}
    x_r^{proj} = x_r^{tar}, \ \ \ \ y_r^{proj} = y_r^{tar},
\end{equation}
where $[x_r, y_r]^T$ represents the 2D root joint. Combined with Eq.~(\ref{equ:p2}), we can get estimated $\hat{X}_r$ and $\hat{Y}_r$,
\begin{equation}
    \hat{X}_r = \frac{\hat{Z}_r(x_r^{tar}-c_x)}{f_x}, \ \ \ \ 
    \hat{Y}_r = \frac{\hat{Z}_r(y_r^{tar}-c_y)}{f_y}.
\end{equation}
Finally, we can obtain the translated root position $\hat{\boldsymbol{J}}_{r} = [\hat{X}_r,\hat{Y}_r,\hat{Z}_r]^T$. The ultimate goal of Global Position Alignment is to align the 2D pose distributions of the target domain and the generated domain (after going through the full data augmentation pipeline) in terms of scale and position. There will be cases of large discrepancies between $J_{3D}^{src}$ and $J_{2D}^{tar}$ when pairing randomly on individuals. However, this is mitigated because 1) we only use box information rather than complete poses and 2) we randomly shuffle the pairing method between each epoch, which further increases diversity and avoids individual discrepancies in a statistical sense.

% We define 2D scale of poses, $\mathcal{S}_{2D}$, as the average of 2D pose width and height
% %, which build the connection between the distributions of 2D pose and 3D pose with the only variable $Z_r$:
% \begin{equation}
% \begin{aligned}
%     \mathcal{S}_{2D} &= \frac{1}{2} \left ( \Delta x^{\prime} + \Delta y^{\prime} \right )\\
%     &= \frac{1}{2} \left [ \Delta \left (\frac{X_i+X_r}{Z_i+Z_r} \right ) + \Delta \left (\frac{Y_i+Y_r}{Z_i+Z_r} \right ) \right ] \\
%     & \approx \frac{1}{2Z_r} \left ( \Delta X + \Delta Y \right )
% \end{aligned}
% \end{equation}
% where $\Delta$ denotes the difference between the max-min coordinates of 2D joints. Therefore the depth $Z_r$ can be estimated by solving the equation:
% \begin{equation}
%     Z_r = \frac{1}{2} \frac{\Delta X+ \Delta Y}{\Delta x + \Delta y}
% \end{equation}

% Finally, applying the projection function~\ref{equ:p} on root joint gives:
% \begin{equation}
% \begin{cases}
%     X_r = x_i (Z_i + Z_r) - X_i = x_i Z_r \\
%     Y_r = y_i (Z_i + Z_r) - Y_i = y_i Z_r
% \end{cases}
% \end{equation}
% since $X_i, Y_i, Z_i = \boldsymbol{0}$ for root joint.

\subsection{Local Pose Augmentation (LPA)}
\label{subsec:ra}

Inspired by \textit{PoseAug}~\cite{gong2021poseaug}, We also apply local pose augmentation (LPA) to enhance the diversity of 2D-3D pose mappings. The augmentation transformation of 3D pose can be decoupled into perturbations of bone vector, bone length, and rotation. We design an adversarial augmentation framework, which consists of an augmentation generator to generate 3D pose transformation parameters and an anchor discriminator to ensure the realisticity, quality, and diversity of generated pose pairs. The generator and discriminator are jointly end-to-end trained following the GAN style.

\noindent\textbf{Augmentation generator}. Following \textit{PoseAug}~\cite{gong2021poseaug}, we propose an augmentation generator denoted as $\mathcal{G}$ with parameters $\boldsymbol{\theta}_\mathcal{G}$. Unlike vanilla GAN-style generator, we take a sample 3D pose from source dataset $J^{src}_{3D}$ as the condition $\mathcal{G}_{cond}$, which is  concatenated with a noise vector $\boldsymbol{z}$ as the input of $\mathcal{G}$, according to the suggestion in~\cite{gholami2022adaptpose, bousmalis2017unsupervised}. The input 3D pose is converted to bone direction vectors representing the joint angle and bone length. Augmentation generator $\mathcal{G}$ generates three types of 3D pose transformations: bone angle difference, bone length difference, and global rotation. The augmentation process can be represented as
\begin{equation}
    \boldsymbol{J}^{aug}_{3D} = \mathcal{G}(\boldsymbol{J}^{src}_{3D}, \boldsymbol{z};\boldsymbol{\theta}_{\mathcal{G}}),\ \ \ \  \boldsymbol{J}^{aug}_{2D} = f_p(\boldsymbol{J}^{aug}_{3D}; \boldsymbol{\theta}_{cam})
\end{equation}
where $(\boldsymbol{J}^{aug}_{3D}, \boldsymbol{J}^{aug}_{2D})$ denotes the augmented paired 2D-3D pose pairs, $f_p$ denotes the camera projection function~\ref{equ:p}, and $\boldsymbol{\theta}_{cam}$ denotes the given camera intrinsic parameters in target dataset.

% \noindent\textbf{Domain Discriminator} Let $\mathcal{D}_{2D}$ denotes the discriminator for 2D pose with parameters $\theta_{\mathcal{D}_{2D}}$. It takes $J_{2D}$ sampled from both $J^{tar}_{2D}$ and $J^{aug}_{2D}$ as input. Discriminator $\mathcal{D}_{2D}$ works as a domain discriminator to distinguish whether the input is from sampled from $J^{tar}_{2D}$ or $J^{aug}_{2D}$.

\noindent\textbf{Anchor discriminator}. Let $\mathcal{D}_{3D}$ denote the discriminator for 3D pose with parameters $\boldsymbol{\theta}_{\mathcal{D}_{3D}}$. It takes root-relative $\boldsymbol{J}_{3D}$ sampled from both $\boldsymbol{J}^{src}_{3D}$ and $\boldsymbol{J}^{aug}_{3D}$ as input. Discriminator $\mathcal{D}_{3D}$ works as an anchor discriminator to ensure the augmented pose $\boldsymbol{J}^{aug}_{3D}$ is reasonable. Inspired by Kinematic Chain Space (KCS)~\cite{wandt2019repnet, wandt2018kinematic}, we use KCS representation of 3D pose as input instead of 3D joints. With the help of KCS representation, the generated 3D poses $\boldsymbol{J}^{aug}_{3D}$ are no longer limited in the source domain.
%Since we assume that there is no access to 3D pose ground truth in target dataset $J^{tar}_{3D}$, we essentially encourage the augmented pose closed to the pose from source dataset in 3D perspective. In this case, discriminator $\mathcal{D}_{3D}$ indeed has a risk of conflict with $\mathcal{D}_{2D}$. However, according to D2GAN~\cite{nguyen2017dual}, we believe that exploiting the complementary statistical properties of two discriminators to improve both the quality and diversity of samples generated from the generator is possible.

\subsection{Training}
\label{subsec:training}
Previous works~\cite{gong2021poseaug, gholami2022adaptpose, yang20183d} train the adversarial pose augmentation framework with the loss function of vanilla GAN~\cite{goodfellow2020generative} or least squares GAN (LSGAN)~\cite{mao2017least}. We argue that training based on Wasserstein distance~\cite{shen2018wasserstein, deng2021synchronized} can be trained stably and provide better augmented poses.

\noindent\textbf{Discriminator loss}. We adopt WGAN~\cite{arjovsky2017wasserstein} loss for the
%discriminator loss $\mathcal{L}_{\mathcal{D}_{2D}}$ and 
anchor discriminator $\mathcal{L}_{\mathcal{D}_{3D}}$:
% \begin{equation}
%     \mathcal{L}_{\mathcal{D}_{2D}} = \mathbb{E} \left [ \mathcal{D}_{2D} (J^{aug}_{2D}) \right]- \mathbb{E} \left [ \mathcal{D}_{2D}(J^{tar}_{2D}) \right ]
% \end{equation}
\begin{equation}
    \mathcal{L}_{\mathcal{D}_{3D}} = \mathbb{E}_{\boldsymbol{x} \sim \boldsymbol{J}^{aug}_{3D}} \left [ \mathcal{D}_{3D}(\boldsymbol{x}) \right]- \mathbb{E}_{\boldsymbol{x} \sim \boldsymbol{J}^{src}_{3D}} \left [ \mathcal{D}_{3D}(\boldsymbol{x}) \right]
\end{equation}
% \begin{equation}
%     \mathcal{L}_{\mathcal{D}_{3D}} = \mathbb{E} \left [ \mathcal{D}_{3D} (J^{aug}_{3D}) \right]- \mathbb{E} \left [ \mathcal{D}_{3D}(J^{src}_{3D}) \right ]
% \end{equation}

\noindent\textbf{Generator loss}. The adversarial loss of 
 the augmentation generator is the feedback from the anchor discriminator.
\begin{equation}
    \mathcal{L}_{\mathcal{G}} = - \mathbb{E}_{\boldsymbol{x} \sim \mathcal{G}(\boldsymbol{J}^{src}_{3D},\boldsymbol{z})} \left [ \mathcal{D}_{3D}(\boldsymbol{x})) \right]
\end{equation}
The discriminator and generator are trained iteratively. 

\noindent\textbf{Lifting network loss}. The standard Mean Squared Error (MSE) loss is adopted to the lifting network $\mathcal{P}$,
\begin{equation}
    \mathcal{L}_{\mathcal{P}} = \Vert \boldsymbol{J}_{GT} - \boldsymbol{J} \Vert_2^2,
\end{equation}
where $\boldsymbol{J}_{GT}$ and $\boldsymbol{J}$ are ground truth and estimated 3D joints, respectively. The generator and the discriminator need to be warmed up for $w$ epoch before training lifting network.

% While training one, the gradient of the other is frozen. % At the beginning of training, we only train the local pose augmentation module to ensure the quality and stability of augmented pose pairs. After 10 epochs for warmup, we train the lifting network with both augmented pose pairs and original pose pairs.

\begin{algorithm}[ht]
\DontPrintSemicolon
  \KwIn{$\boldsymbol{J}=\{\boldsymbol{J}^{src}_{3D},\boldsymbol{J}^{tar}_{2D},\boldsymbol{\theta}_{cam}\}$}
  \BlankLine
  \For{$t\leftarrow 1$ \KwTo $T$}{
    \For{$i\leftarrow 1$ \KwTo $I$}{
        \textbf{freeze} $\mathcal{G}$\\
        \emph{sample and generate a batch data}\\
        $\boldsymbol{J}^{aug}_{3D}, \boldsymbol{J}^{aug}_{2D} \gets \mathcal{G}(\boldsymbol{J}^{src}_{3D}, \boldsymbol{z}; \boldsymbol{\theta}_{\mathcal{G}})$\\
        \emph{train discriminator}\\
        $\mathcal{L}_{\mathcal{D}_{3D}} = \mathbb{E}_{\boldsymbol{x} \sim \boldsymbol{J}^{aug}_{3D}} \left [ \mathcal{D}_{3D}(\boldsymbol{x}; \boldsymbol{\theta}_{\mathcal{D}_{3D}}) \right]- \mathbb{E}_{\boldsymbol{x} \sim \boldsymbol{J}^{src}_{3D}} \left [ \mathcal{D}_{3D}(\boldsymbol{x}; \boldsymbol{\theta}_{\mathcal{D}_{3D}}) \right]$\\
        \textbf{update} $\boldsymbol{\theta}_{\mathcal{D}_{3D}}$\\
        \emph{train generator every $n$ iters}\\
        \If{$i$ in every $n$ iters}{
            \textbf{freeze} $\boldsymbol{\theta}_{\mathcal{D}_{3D}}$\\
            \emph{generate a batch data}
            $\boldsymbol{J}^{aug}_{3D}, \boldsymbol{J}^{aug}_{2D} \gets \mathcal{G}(\boldsymbol{J}^{src}_{3D}, \boldsymbol{z}; \boldsymbol{\theta}_{\mathcal{G}})$\\
            $\mathcal{L}_{\mathcal{G}} = - \mathbb{E}_{\boldsymbol{x} \sim \boldsymbol{J}^{aug}_{3D}} \left [ \mathcal{D}_{3D}(\boldsymbol{x}; \boldsymbol{\theta}_{\mathcal{D}_{3D}}) \right]$\\
            \textbf{update} $\boldsymbol{\theta}_{\mathcal{G}}$\\
        }
    }
    \emph{warmup to ensure stable augmentation}\\
    \If{$t\geq w$}{
        \emph{mixup augmented and source data}\\
        $\boldsymbol{J}^{mix}_{2D}, \boldsymbol{J}^{mix}_{3D} \gets \left ( \boldsymbol{J}^{src}_{2D}, \boldsymbol{J}^{src}_{3D}\right ), \left ( \boldsymbol{J}^{aug}_{2D}, \boldsymbol{J}^{aug}_{3D}\right )$\\
        \emph{train lifting network with mixed data}\\
        $\mathcal{L}_{\mathcal{P}} = \text{MSE} \left (\mathcal{P}\left( \boldsymbol{J}^{mix}_{2D};\boldsymbol{\theta}_{\mathcal{P}}\right), \boldsymbol{J}^{mix}_{3D} \right)$\\
        \textbf{update} $\boldsymbol{\theta}_{\mathcal{P}}$\\
    }
}

\caption{The training pipeline of our method}
\label{training pipeline}
\end{algorithm}
\section{Experiments}
\label{exp}

In this section, we conduct experiments on several popular human pose estimation benchmarks with comprehensive evaluation metrics. Since \textit{PoseDA} is flexible regarding the architecture of the pose lifting network, we perform our framework on different strong baselines to show the universality. GT 2D pose is used as default. We also analyze the contribution of each component in ablation studies.

\begin{figure}
    \centering
    \includegraphics[width=1\linewidth]{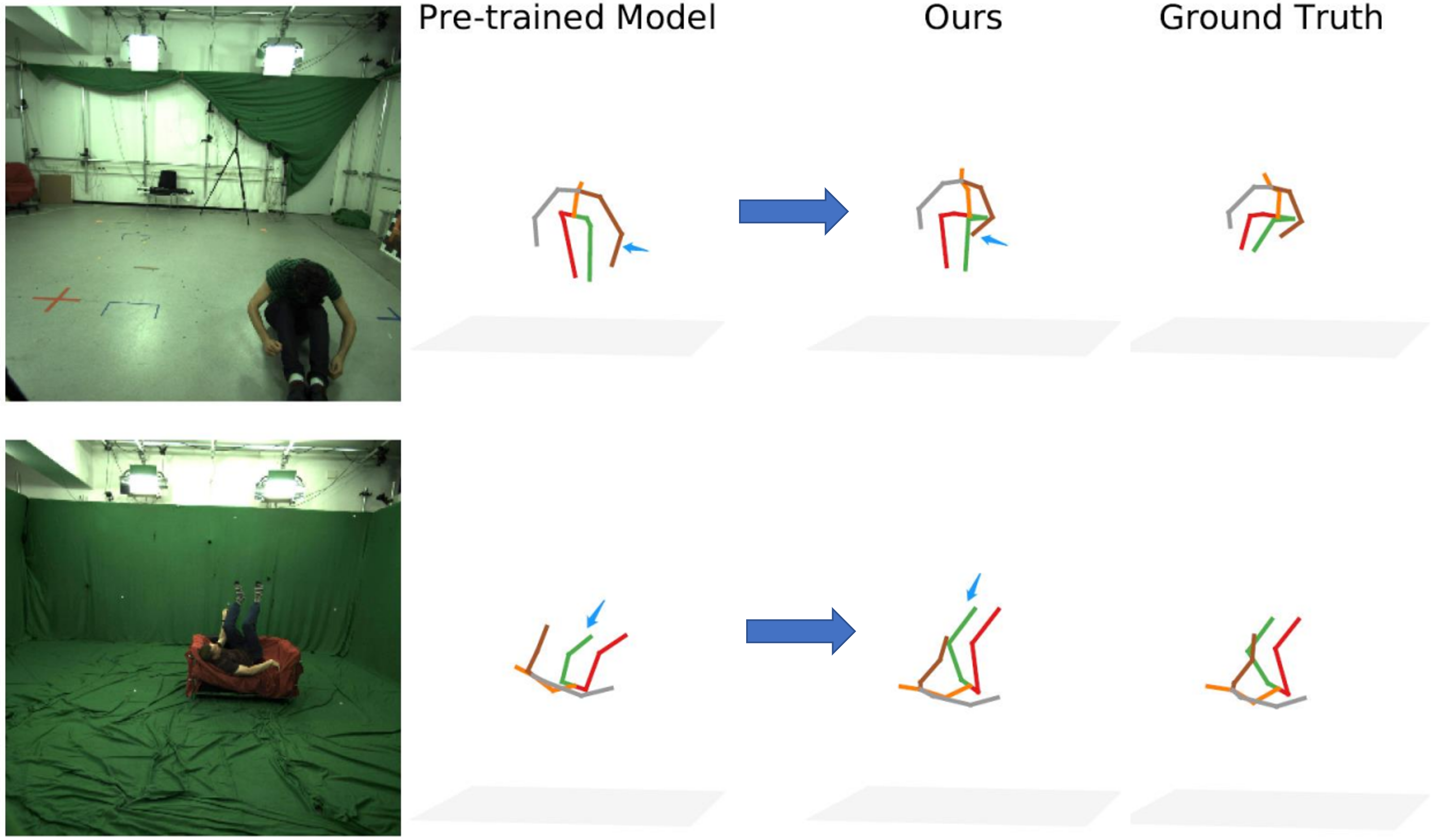}
    \caption{\small Qualitative results on MPI-INF-3DHP dataset. We use VideoPose3D~\cite{pavllo20193d} as pre-trained model. After finetuning with \textit{PoseDA}, the model performs better for some challenging poses.}
    \label{fig:qualitative}
    % \vspace{-1em}
\end{figure}

\subsection{Datasets and Metrics}
\label{subsec:datasets}

The datasets used for quantitative evaluations are several widely used large-scale 3D human pose estimation benchmarks, including Human3.6M~\cite{ionescu2013human3}, MPI-INF-3DHP~\cite{mehta2017monocular}, and 3DPW~\cite{von2018recovering}.

\noindent\textbf{Human3.6M (H3.6M)} is one of the largest 3D human pose datasets captured in a laboratorial environment. Following previous works~\cite{gholami2022adaptpose, pavllo20193d}, there are two different settings of H3.6M: 1) using the S1, S5, S6, S7 and S8 as our source domain for cross-dataset evaluation. 2) using only S1 as source domain and others (S5, S6, S7, S8) as target domain.

\noindent\textbf{MPI-INF-3DHP (3DHP)} is a large-scale in-the-wild 3D human pose dataset with more diverse actions and motions. This dataset is closer to real-world scenarios and ideal for evaluating our method. Following previous works~\cite{kolotouros2019learning, gong2021poseaug}, we use its test set, which includes 2,929 frames. We report the results of \textit{PoseDA} using metrics of MPJPE, Percentage of Correct Keypoints (PCK), and Area Under the Curve (AUC).

\noindent\textbf{3DPW} is an in-the-wild dataset, unlike H3.6M or 3DHP, with uncontrolled motion and scene. Since it is a much more challenging dataset, we train each method on H3.6M and evaluate it on the 3DPW test set with the PA-MPJPE and MPJPE metric. 

\begin{table}[!t]
\small
\centering
\setlength{\tabcolsep}{1mm}

\begin{tabular}{l|c|cc}
    \specialrule{1pt}{1pt}{2pt}
    % \toprule
    Method & S & MPJPE~($\downarrow$) & PA-MPJPE~($\downarrow$)\\
    \hline
    VideoPose3D~\cite{pavllo20193d} & Full & 51.8 & 40.0 \\
    ST-GCN~\cite{cai2019exploiting} & Full & 50.6 &  40.2\\
    SimpleBaseline~\cite{martinez2017simple} & Full & 45.5 & 37.1 \\
    SemGCN~\cite{zhao2019semantic} & Full & 43.8 & -\\
    \hline
    VideoPose3D~\cite{pavllo20193d} & S1 & 64.7 & -\\
    Li~\etal~\cite{li2020cascaded} & S1 & 62.9 & - \\
    PoseAug~\cite{gong2021poseaug} & S1 & 56.7 & -\\
    AdaptPose~\cite{gholami2022adaptpose} & S1 & 54.2 & 35.6\\
    %AdaptPose~(T=27) & S1 & 42.5 & 34.0\\
    \hline
    \textit{PoseDA} (Ours)  & S1 & \textbf{49.9} & \textbf{34.2}\\
    % \bottomrule
    \specialrule{1pt}{1pt}{2pt}
\end{tabular}
\caption{\small \textbf{Results on H3.6M}. S denotes source data. MPJPE and PA-MPJPE are used as  evaluation metrics. Source: H3.6M-S1. Target: H3.6M-S5, S6, S7, S8.}
\label{tab:H3.6M}
\end{table}

\subsection{Implementation Details}

\noindent\textbf{Details of discriminators}.
For the anchor discriminator, we split 3D keypoints into five parts following~\cite{gong2021poseaug}, pass them through each of the five 4-layer residual MLPs with LeakyReLU, and finally concatenate the output. We use RMSProp optimizer for the anchor discriminators.

\noindent\textbf{Details of generators}.
We sequentially build three 3-layer residual MLPs with LeakyReLU to generate bone angle, bone length, and rotation. Each network receives the output of the previous network as well as noise vector as input. These three networks are trained jointly by weighted adversarial losses. For every $6$ iterations, we train the generator once and the discriminator $5$ times. 

\noindent\textbf{Details of pose lifting network}.
We use VideoPose3D~\cite{pavllo20193d} (1-frame) as pose lifting network as well as other strong baselines~\cite{cai2019exploiting, martinez2017simple, zhao2019semantic} in ablation study. The pre-trained weights from the source dataset are used as initial weights for all the experiments.

\noindent\textbf{Details of training}.
Starting with pre-trained weight, it takes 5 to 10 epochs (50 secs per epoch) to get convergence. In the experiments, we also found that WGAN not only has more stable training than LSGAN, but also prevents mode collapse, but the performance is not much different.We only train the generator and discriminator at the first $5$ epochs for stable augmentation. Then, we mix the augmented data and original data in an 1:1 ratio for pose lifting network training. The learning rate for all the networks is $1e^{-4}$. \textit{PoseDA} is trained on NVIDIA RTX 3090. The training process takes two hours and consumes 2GB of GPU memory. 

\subsection{Results}
\label{subsec:results}

% \begin{table}[h] \centering
% %\ra{1.3}
% \caption{Comparison of 3D human pose estimation datasets we used for training and evaluating. The mean and std of camera distance, camera height, focal length, bone length is calculated in \cite{wang2020predicting}. Focal length is in mm while the others are in unit meters. We use the camera intrinsics for target dataset while training. We also align the number and definition of joints in different datasets following~\cite{gong2021poseaug}.}
% \begin{center}
% {\scriptsize
% \vspace{-0.5cm}
% \begin{tabular}{@{}l c c c@{}}
% %\Xhline{2\arrayrulewidth}
% \toprule
% Dataset & H3.6M & 3DPW & 3DHP \\
% \hline
% Year & 2014  & 2018 & 2017 \\
% Imaging Space & 1000 $\times$ 1002  & 1920 $\times$ 1080 & 2048 $\times$ 2048 \\ 
% Camera Distance & 5.2 $\pm$ 0.8 & 3.5 $\pm$ 0.7 & 3.8 $\pm$ 0.8 \\
% Camera Height & 1.6 $\pm$ 0.05  & 0.6 $\pm$ 0.8 & 0.8 $\pm$ 0.4 \\
% Focal Length& 1146.8 $\pm$ 2.0  & 1962.2 $\pm$ 1.5  & 1497.88 $\pm$ 2.8 \\
% %Center & H3.6M & gpa & surreal & 3dpw & 3dhp \\
% No. of Joints & 38 & 24 & 28 or 17 \\
% No. of Cameras & 4 & 1 & 14 \\
% No. of Subjects & 11 & 18 & 8 \\
% Bone Length & 3.9 $\pm$ 0.1  & 3.7 $\pm$ 0.1 & 3.7 $\pm$ 0.1 \\
% GT source &VICON &  SMPL & The Capture \\
% No. Train Images & 311,951 & 22,375 & 366,997 \\
% No. Test Images  & 109,764 & 35,515 &  2,929\\
% Scene & controlled indoor & natural outdoor & controlled in/outdoor \\
% %\Xhline{2\arrayrulewidth}
% \bottomrule
% \end{tabular}
% }
% \end{center}
% \vspace{-0.5cm}
% \label{tab:datasets}
% \end{table}

\noindent\textbf{Results on H3.6M}. We compare \textit{PoseDA} with state-of-the-art methods~\cite{pavllo20193d,li2020cascaded,gong2021poseaug,gholami2022adaptpose} using ground truth 2D keypoints as input. The results on H3.6M are shown in Table~\ref{tab:H3.6M}, focusing on the generalization ability of the model over actions, since the distributions over global positions are relatively consistent between the different parts of the same dataset. Even though AdaptPose~\cite{gholami2022adaptpose} utilizes temporal information, our method still achieves state-of-the-art performance.

\begin{table}[!t]
\small
\centering
\setlength{\tabcolsep}{1mm}

\begin{tabular}{l|c|ccc}
    \specialrule{1pt}{1pt}{2pt}
    Method & CD & MPJPE~($\downarrow$) & PCK~($\uparrow$) & AUC~($\uparrow$) \\
    \hline
    Multi Person~\cite{mehta2018single} & & 122.2 & 75.2 & 37.8 \\
    Mehta~\etal~\cite{mehta2017monocular} & & 117.6 & 76.5 & 40.8 \\
    VNect~\cite{mehta2017vnect}  & & 124.7 & 76.6 & 40.4 \\
    OriNet~\cite{luo2018orinet} & & 89.4 & 81.8 & 45.2  \\
    SimpleBaseline~\cite{martinez2017simple} & & 84.3 & 85.0 & 52.0\\
    % Chen~\etal(T=81)~\cite{chen2021anatomy} & & 78.8 & 87.9 & 54.0\\
    % PoseFormer(T=9)~\cite{zheng20213d} & & 77.1 & 88.6 & 56.4\\
    % CrossFormer(T=9)~\cite{hassanin2022crossformer} & & 76.3 & 89.1 & 57.5\\
    % MHFormer(T=9)~\cite{li2022mhformer} & & 58.0 & 93.8 & 63.3 \\
    MixSTE~\cite{zhang2022mixste} & & 57.9 & 94.2 & 63.8 \\
    \hline	
    LCN~\cite{ci2019optimizing} & \checkmark & - & 74.0 & 36.7\\
    HMR~\cite{kanazawa2018end} & \checkmark & 113.2 & 77.1 & 40.7\\
    SRNet~\cite{zeng2020srnet} & \checkmark & - & 77.6 & 43.8\\
    Li~\textit{et al.}~\cite{li2020cascaded} & \checkmark & 99.7 & 81.2 & 46.1 \\
    RepNet~\cite{wandt2019repnet} & \checkmark & 92.5 & 81.8 & 54.8 \\
    PoseAug~\cite{gong2021poseaug}  & \checkmark & 73.0 & 88.6 & 57.3\\
    AdaptPose~\cite{gholami2022adaptpose} & \checkmark & 68.3 & 90.2 & 59.0\\
    \hline
    \textit{PoseDA} (Ours)  & \checkmark & \textbf{61.3}& \textbf{92.1} & \textbf{62.5} \\
    \specialrule{1pt}{1pt}{2pt}
\end{tabular}
\caption{\small \textbf{Results on 3DHP}. CD denotes cross-domain evaluation (no CD denotes fully-supervision, i.e., trained and tested on the same 3DHP dataset). PCK, AUC and MPJPE are used as evaluation metrics. Source: H3.6M. Target: 3DHP.}
\label{tab:3dhp}
\end{table}

\noindent\textbf{Results on 3DHP}. Our method achieves remarkable performance in all the metrics. We use single frame ground truth 2D keypoints as input and therefore compare against various recent methods with the same setting. \textit{PoseDA} improve upon the previous state-of-the-art method by 10.2\%, and is even competitively compared with some state-of-the-art fully supervised method.

\noindent\textbf{Results on 3DPW}~\cite{von2018recovering}. We use ground truth 2D keypoints as input. \textit{PoseDA} achieves the state-of-the-art performance without using any 3D annotations in 3DPW, as shown in Table~\ref{tab:3dpw}, even is favorably compared with video-based methods.

\begin{table}[!t]
\small
\centering
\setlength{\tabcolsep}{1mm}

\begin{tabular}{l|c|c|cc}
    \specialrule{1pt}{1pt}{2pt}
    Method & CD & V & PA-MPJPE~($\downarrow$) & MPJPE~($\downarrow$)\\
    \hline
    VideoPose3D~\cite{pavllo20193d} & &\checkmark & 68.0 & 105.0\\
    EFT~\cite{joo2021exemplar} & & & 55.7 & -\\
    VIBE~\cite{kocabas2020vibe} & &\checkmark & 51.9 & 82.9\\
    Lin~\etal~\cite{lin2021mesh} & & & 45.6 & 74.7\\
    \hline
    PoseAug~\cite{gong2021poseaug} & \checkmark& & 58.5 & 94.1\\
    VIBE~\cite{kocabas2020vibe} & \checkmark&\checkmark & 56.5 & 93.5\\
    BOA~\cite{guan2021bilevel} & \checkmark&\checkmark & 49.5 & \textbf{77.2}\\
    AdaptPose~\cite{gholami2022adaptpose} &\checkmark& \checkmark &\textbf{46.5} & 81.2\\
    \hline
    \textit{PoseDA} (Ours) & \checkmark && 55.3 & 87.7\\
    \specialrule{1pt}{1pt}{2pt}
\end{tabular}
\caption{\small \textbf{Results on 3DPW}. CD denotes cross-domain evaluation (no CD denotes fully-supervision, i.e., trained and tested on the same 3DPW dataset), V denotes video-based method. PA-MPJPE and MPJPE are used as evaluation metrics. Source: H3.6M. Target: 3DPW.}
\label{tab:3dpw}
\end{table}

\noindent\textbf{Qualitative evaluation}. Figure~\ref{fig:qualitative} shows the qualitative evaluation on 3DHP dataset. Compared with the baseline model without training with \textit{PoseDA}, our method performs well for unusual human positions and challenging poses.

\begin{table}[!t]
    \small
    \centering
    \newcommand{\D}[1]{~\scriptsize{\textcolor{red}{(#1)}}}
    \setlength{\tabcolsep}{1mm}
    
    \begin{tabular}{l|ccc}
        \specialrule{1pt}{1pt}{2pt}
        Method & MPJPE~($\downarrow$)  & PCK~($\uparrow$) & AUC~($\uparrow$)\\
        \hline
        SemGCN~\cite{zhao2019semantic} & 95.96 & 80.68 & 48.48 \\
        + LPA & 87.64\D{-8.3} & 84.21\D{+3.5} & 51.24\D{+2.8}\\
        + GPA & 86.56\D{-9.4} & 83.85\D{+3.2} & 50.98\D{+2.5} \\
        \rowcolor[gray]{0.9}
        + \textit{PoseDA}& \textbf{78.37}\D{-17.6} & \textbf{86.17}\D{+5.5} & \textbf{54.74}\D{+6.3}\\
        \hline
        SimpleBaseline~\cite{martinez2017simple} & 81.23 & 85.85  & 53.95\\
        + LPA & 66.56\D{-14.7} & 90.16\D{+4.3} & 60.41\D{+6.5} \\
        + GPA & 69.19\D{-12.0} & 89.90\D{+4.1} & 58.50\D{+4.6}\\
        \rowcolor[gray]{0.9}
        + \textit{PoseDA}& \textbf{64.79}\D{-16.4} & \textbf{90.55}\D{+4.7} & \textbf{61.32}\D{+7.4}\\
        \hline
        ST-GCN~\cite{cai2019exploiting} & 81.19 & 85.92 & 53.78\\
        + LPA & 74.31\D{-6.9} & 88.72\D{+2.9} & 56.20\D{+2.4}\\
        + GPA & 74.41\D{-6.8} & 88.58\D{+2.7} & 55.52\D{+1.7}\\
        \rowcolor[gray]{0.9}
        + \textit{PoseDA}& \textbf{69.50}\D{-11.7} & \textbf{90.15}\D{+4.2} & \textbf{58.56}\D{+4.8}\\
        \hline
        VideoPose3D~\cite{pavllo20193d} & 82.55 & 85.71 & 53.35\\
        + LPA & 66.65\D{-15.9} & 90.05\D{+4.3} & 60.24\D{+6.9}\\
        + GPA & 66.07\D{-16.5} & 90.87\D{+5.2} & 60.07\D{+6.7}\\
        \rowcolor[gray]{0.9}
        + \textit{PoseDA}&  \textbf{61.36}\D{-21.2}& \textbf{92.05}\D{+6.3} & \textbf{62.52}\D{+9.2} \\
        \specialrule{1pt}{1pt}{2pt}
    \end{tabular}
    \caption{\small Ablation study on components and pose lifting network of our method. LPA denotes local pose augmentation, GPA denotes global position alignment. \textit{PoseDA} combine GPA and LPA in a unified framework. Source: H3.6M. Target: 3DHP.}
    \label{tab:ab1}
\end{table}

\subsection{Ablation Study}
\label{ablation}
    
\noindent\textbf{Analysis on each components and board lifting network}.
Since \textit{PoseDA} is a data augmentation framework, with any pose lifting network architecture. Shown in Tabel~\ref{tab:ab1}, we conduct experiments on several strong baselines with different architectures, including MLP~\cite{martinez2017simple}, Convolutional Network~\cite{pavllo20193d}, and Graph Convolutional Network~\cite{cai2019exploiting, zhao2019semantic}. We also check the effectiveness of each component in \textit{PoseDA}. For fair comparison, we use the same weights of generator and discriminators in all the experiments. It shows that both global position alignment (GPA) and local pose augmentation (LPA) benefit the adaptive performance. Moreover, GPA significantly boosts performance without any extra training or use of additional learnable parameters.

% \noindent\textbf{Analysis on global position alignment (GPA)}
% We show that global position alignment module actually align the 2D pose distribution in terms of scale and root position on 2D screen in Figure~\ref{fig:position}.

% \begin{figure}[h]
%     \centering
%     \includegraphics[width=0.95\linewidth]{figures/scale.pdf}
    
%     \includegraphics[width=0.95\linewidth]{figures/x.pdf}

%     \includegraphics[width=0.95\linewidth]{figures/y.pdf}

%     \caption{Comparison of 2D scale (first row), root position of x-axis (second row), y-axis (third row) in source domain (left), source domain after GPA (middle), target domain (right). Source: H3.6M. Target: 3DHP.}

%     \label{fig:position}
% \end{figure}

\begin{figure}[!t]
    \centering
    \includegraphics[width=0.9\linewidth]{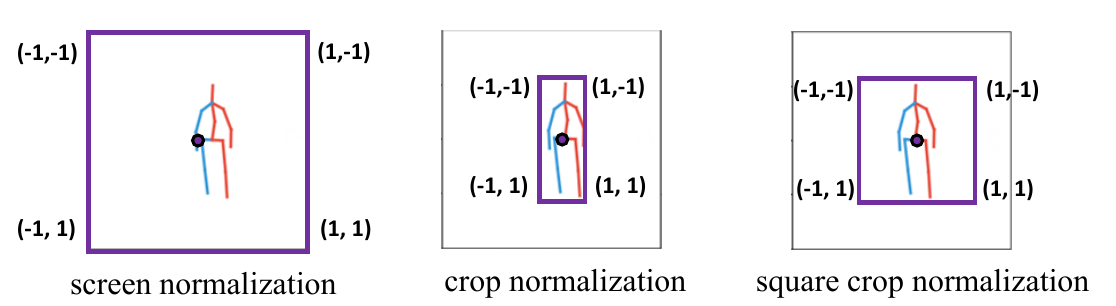}
    \caption{\small Illustration of different types of normalization operations.}
    \label{fig:my_label}
\end{figure}

\noindent\textbf{Analysis on global pose alignment (GPA)}. We argue that the crop and normalization on 2D poses is an inaccurate method compared to our GPA module. Because the projection relation is not a linear operator and therefore does not have translation invariance. The crop out loses information about the relative positions of the camera and the person, and when the same 2d pose is cropped in a different position in the picture, the corresponding 3d pose is different (even if it is root-relative), there will be a one-to-many situation, which still increases ambiguity. We conduct experiments under four different pre-processing settings and with two backbones, our GPA based on screen normalized. We train the model on Human3.6M and test on 3DHP with the same pre-processing. Figure \ref{fig:my_label} shows different normalization operations and Table~\ref{tab:ab1} shows our GPA method outperforms other normalization methods.

% 我们相信crop和normalization的方法相比于我们的GPA是有误差的方法，这是因为投影关系不是一个线性算子，因此不具备平移不变性。[Optional] 以下是数学证明
% Let $S=([\boldsymbol{R}_{2D}, \boldsymbol{J}^i_{2D}], [\boldsymbol{R}_{3D}, \boldsymbol{J}^i_{3D}])$ denote corresponding 2D-3D pose pairs, where $\boldsymbol{R}_{2D}=[x_r, y_r]$, $\boldsymbol{R}_{3D}=[X_r, Y_r, Z_r]$ is the root position and $\boldsymbol{J}^i_{2D}=[x_i, y_i]$, $\boldsymbol{J}^i_{3D}=[X_i, Y_i, Z_i]$ is the position of $i$-th joint. We define \textbf{crop} operation to be equivalent to setting $\boldsymbol{R}_{2D}$ to 0 and \textbf{normalization} opeartion to be equivalent to multiplying a factor $s$ on $\boldsymbol{J}^i_{2D}$. The projection from 3D joints to 2D joints after translation should obey the perspective projection function as follows. without loss of generality, we only discuss the x-axis:
% \begin{equation}
%     \frac{f_x(X_i+X_r)}{Z_i+Z_r}+c_x = x_i+x_r
% \end{equation}
% \vspace{-0.4cm}
% \vspace{-0.6cm}

% [ADD] 实验结论 GPA use screen normalized 2D coordinate

\begin{table}[!t]
    \small
    \centering
    \newcommand{\D}[1]{~\scriptsize{\textcolor{red}{(#1)}}}
    \newcommand{\U}[1]{~\scriptsize{\textcolor{green}{(#1)}}}
    \setlength{\tabcolsep}{1mm}
    
    \begin{tabular}{l|ccc}
        \specialrule{1pt}{1pt}{2pt}
        Method & MPJPE~($\downarrow$)  & PCK~($\uparrow$) & AUC~($\uparrow$)\\
        \hline
        SimpleBaseline \\
        w screen norm & 81.23 & 85.85 & 53.95\\
        w crop norm & 95.37 & 81.40 & 47.66\\
        w square crop & 84.81 & 84.29 & 52.27\\
        \rowcolor[gray]{0.9}
        w GPA (\textit{ours}) & \textbf{69.19} & \textbf{89.90} & \textbf{58.50}\\
        \hline
        VideoPose3D\\
        w screen norm & 82.55 & 85.71 & 53.35\\
        w crop norm & 93.16 & 82.23 & 48.00\\
        w square crop & 85.05 & 84.90 & 51.81\\
        \rowcolor[gray]{0.9}
        w GPA (\textit{ours}) & \textbf{66.07} & \textbf{90.87} & \textbf{60.07}\\
        \specialrule{1pt}{1pt}{2pt}
    \end{tabular}
    % \vspace{-0.2cm}
    \caption{\small Ablation study on Global Position Alignment over other normalization techniques. Source: H3.6M. Target: 3DHP.}
    \label{tab:ab1}
\end{table}

\begin{table}[!t]
    \small
    \centering
    \setlength{\tabcolsep}{1mm}
    
    \begin{tabular}{l|ccc}
        \specialrule{1pt}{1pt}{2pt}
        Method & MPJPE~($\downarrow$)  & PCK~($\uparrow$) & AUC~($\uparrow$)\\
        \hline
        No Aug & 66.07 & 90.87 & 60.07\\
        RR & 66.88 & 90.77 & 59.31\\
        BL & 65.81 & 90.94 & 59.91\\
        RR + BL & 64.99 & 91.07 & 60.49\\
        \textit{PoseDA} & \textbf{61.36} & \textbf{92.05} & \textbf{62.52}\\
        \specialrule{1pt}{1pt}{2pt}
    \end{tabular}
    \caption{\small Ablation study on local pose augmentation with different pre-defined pose augmentation methods. RR denotes random rotation along vertical axial and BL denotes random bone length transformation. Source: H3.6M. Target: 3DHP.}
    \label{tab:ra}
\end{table}

\noindent\textbf{Analysis on local pose augmentation (LPA)}. We also conduct ablation study on  different methods of LPA. Comparing with random rotation and random bone length transformation, the proposed adversarial augmentation method achieves better performance as shown in Table~\ref{tab:ra}.

\subsection{Discussion Compared to SOTA Methods}

% \begin{figure*}[!t]
%     \centering
%     \includegraphics[width=0.95\linewidth]{figures/dis.pdf}
%     \caption{Discussion of input selection of generator condition, real 3d pose, and real 2d pose in local pose augmentation (LPA). Real pose means the goal of the discriminator. We show that applying adaptation on local pose actually degrades the performance.}
%     \vspace{-1.0em}
%     \label{fig:dis}
% \end{figure*}

The most related SOTA works are PoseAug and AdaptPose. Our proposed GPA has significant differences from the two methods in several aspects. In addition, our design of the discriminator is also different from these two methods.
PoseAug is a simple domain generalization framework that uses two discriminators, called $D_{3D}$ and $D_{2D}$. Their goal is to regulate the poses generated by the generator, making them similar to the 3D and 2D poses of the source domain. AdaptPose is a domain adaptation framework that also uses these two discriminators, but the goal of $D_{2D}$ is to make the generated poses more similar to the 2D poses of the target domain. However, the assumption of AdaptPose is not supported after careful and comprehensive experiments. In the human pose domain adaptation problem, we abandon $D_{2D}$ and only use $D_{3D}$ to regulate the quality of the generated poses to achieve better results. The reason we give is that: 1) Compared to PoseAug, making the generated 2D poses unnecessary similar to the source domain brings greater diversity. 2) Compared to AdaptPose, we believe that forcing the generated 2D poses to be similar to the target domain does not guarantee that the corresponding 3D poses are also similar. In other words, this is also because the 2D-3D mapping has large ambiguity. As a result, with our carefully designed GPA and LPA modules, global adaptation and local generalization are well combined, which is also our major contribution.

% As for why AdaptPose is better than PoseAug, we think the translation part in the generator plays a role, and forms a transfer on the projected 2D scale and position. While other works use camera view estimation or generation as an auxiliary task to address the global position adaptation problems, GPA achieves alignment explicitly and directly in this part.

We conduct extensive and convincing experiments on the input selection of local pose augmentation module. All the experiments in this section are based on pre-trained VideoPose3D and global position alignment. Following~\cite{gholami2022adaptpose, gong2021poseaug}, we design a 2D pose discriminator of 5-layer MLP. Note that the condition of generator is used only to generate transformation, we still apply those transformation on the 3D pose from the source dataset. The 3D pose of target dataset used in 3D discriminator is the prediction of corresponding 2D pose by the pre-trained lifting network. As shown in Table~\ref{tab:dis}, these experiments lead to two important conclusions: 1) the design of 2D pose discriminator is unnecessary, and 2) there is a performance drop no matter either the target domain information is involved in the generator or the discriminator to adapt the characteristics of local pose. 

% We argue that the adaption on the local pose in 2D (\textit{e.g.}, bone vector) actually \textit{hurts} the diversity of the 2D-3D mapping. To be specific, the human with the same 2D pose may act totally different in 3D space. The alignment of 2D local pose actually means nothing in 3D. According to this viewpoint, PoseAug~\cite{gong2021poseaug} does data augmentation on both global position and local pose, and AdaptPose~\cite{gholami2022adaptpose} focuses on domain adaption on both of them. In our case, we do domain adaptation on global position and augmentation on local pose, achieving the best performance.

\begin{table}[!t]
    \small
    \centering
    \setlength{\tabcolsep}{1mm}
    \begin{tabular}{ccc|ccc}
        \specialrule{1pt}{1pt}{2pt}
        $\mathcal{G}_{cond}$ & $\mathcal{D}_{3D}$ & $\mathcal{D}_{2D}$ & MPJPE~($\downarrow$) & PCK~($\uparrow$) & AUC~($\uparrow$)\\
        \hline
        - & - & - & 66.07 & 90.87 & 60.07\\
        \hline
        $\mathcal{T}$ & $\mathcal{T}$ & $\mathcal{T}$ & 66.34 & 90.81 & 59.72\\
        $\mathcal{T}$ & $\mathcal{S}$ & $\mathcal{T}$ & 65.91 & 91.02 & 59.92\\
        $\mathcal{T}$ & $\mathcal{S}$ & - & 65.37 & 90.98 & 60.30\\
        \hline
        $\mathcal{S}$ & $\mathcal{T}$ & $\mathcal{T}$ & 64.20 & 91.64 & 60.85\\
        $\mathcal{S}$ & $\mathcal{T}$ & - & 63.66 & 91.57 & 61.34\\
        \hline
        $\mathcal{S}$ & $\mathcal{S}$ & $\mathcal{S}$ & 73.55 & 88.96 & 56.41\\
        $\mathcal{S}$ & $\mathcal{S}$ & $\mathcal{T}$ & 65.46 & 91.27 & 60.03\\
        $\mathcal{S}$ & $\mathcal{S}$ & - & \textbf{61.36}& \textbf{92.05} & \textbf{62.52}\\
        \specialrule{1pt}{1pt}{2pt}
    \end{tabular}
    \caption{\small The input selection of generator condition $\mathcal{G}_{cond}$, 3D pose discriminator $\mathcal{D}_{3D}$, and 2D pose discriminator $\mathcal{D}_{2D}$ in LPA. $S, T$ denote the pose from source or target domain. Source: H3.6M. Target: 3DHP.}
    \label{tab:dis}
\end{table}   

\section{Conclusion}
\label{conclusion}

This paper addresses the problem of unsupervised cross-domain adaptation for 3D human pose estimation. To reduce the domain gap, we propose global position alignment and local pose augmentation. We argue that global position alignment is simple yet effective, and the local pose augmentation enhances the diversity of 2D-3D pose mapping. The proposed global position alignment module significantly boosts performance with no additional learnable parameters needed. We also show that adversarial pose augmentation based on Wasserstein distance can further obtain stable, diverse, and high-quality pose pairs. With extensive and convincing experiments and ablation studies, \textit{PoseDA} can be flexibly applied on any 2D-3D pose lifting network and make a significant step towards solving domain adaptation problems for 3D human pose estimation.

\paragraph{Limitations and Future Work} Although we show that generalization performs better than adaptation on domain gap over local pose, it is still possible to design a method that can adapt 3D pose without hurting the diversity of 2D-3D mappings.

\paragraph{Acknowledgement} This work is supported by the National Key R\&D Program of China No.2022ZD0162000, and National Natural Science Foundation of China No.62106219.

%%%%%%%%% REFERENCES
{\small
\bibliographystyle{ieee_fullname}
\bibliography{ref}

\begin{thebibliography}{10}\itemsep=-1pt

\bibitem{andriluka2018posetrack}
Mykhaylo Andriluka, Umar Iqbal, Eldar Insafutdinov, Leonid Pishchulin, Anton
  Milan, Juergen Gall, and Bernt Schiele.
\newblock Posetrack: A benchmark for human pose estimation and tracking.
\newblock In {\em Proceedings of the IEEE conference on computer vision and
  pattern recognition}, pages 5167--5176, 2018.

\bibitem{arjovsky2017wasserstein}
Martin Arjovsky, Soumith Chintala, and L{\'e}on Bottou.
\newblock Wasserstein generative adversarial networks.
\newblock In {\em International conference on machine learning}, pages
  214--223. PMLR, 2017.

\bibitem{bousmalis2017unsupervised}
Konstantinos Bousmalis, Nathan Silberman, David Dohan, Dumitru Erhan, and Dilip
  Krishnan.
\newblock Unsupervised pixel-level domain adaptation with generative
  adversarial networks.
\newblock In {\em Proceedings of the IEEE conference on computer vision and
  pattern recognition}, pages 3722--3731, 2017.

\bibitem{cai2019exploiting}
Yujun Cai, Liuhao Ge, Jun Liu, Jianfei Cai, Tat-Jen Cham, Junsong Yuan, and
  Nadia~Magnenat Thalmann.
\newblock Exploiting spatial-temporal relationships for 3d pose estimation via
  graph convolutional networks.
\newblock In {\em Proceedings of the IEEE/CVF international conference on
  computer vision}, pages 2272--2281, 2019.

\bibitem{chen20173d}
Ching-Hang Chen and Deva Ramanan.
\newblock 3d human pose estimation= 2d pose estimation+ matching.
\newblock In {\em Proceedings of the IEEE conference on computer vision and
  pattern recognition}, pages 7035--7043, 2017.

\bibitem{chen2019unsupervised}
Ching-Hang Chen, Ambrish Tyagi, Amit Agrawal, Dylan Drover, Rohith Mv, Stefan
  Stojanov, and James~M Rehg.
\newblock Unsupervised 3d pose estimation with geometric self-supervision.
\newblock In {\em Proceedings of the IEEE/CVF Conference on Computer Vision and
  Pattern Recognition}, pages 5714--5724, 2019.

\bibitem{chen2022epro}
Hansheng Chen, Pichao Wang, Fan Wang, Wei Tian, Lu Xiong, and Hao Li.
\newblock Epro-pnp: Generalized end-to-end probabilistic perspective-n-points
  for monocular object pose estimation.
\newblock In {\em Proceedings of the IEEE/CVF Conference on Computer Vision and
  Pattern Recognition}, pages 2781--2790, 2022.

\bibitem{chen2016synthesizing}
Wenzheng Chen, Huan Wang, Yangyan Li, Hao Su, Zhenhua Wang, Changhe Tu, Dani
  Lischinski, Daniel Cohen-Or, and Baoquan Chen.
\newblock Synthesizing training images for boosting human 3d pose estimation.
\newblock In {\em 2016 Fourth International Conference on 3D Vision (3DV)},
  pages 479--488. IEEE, 2016.

\bibitem{chen2018cascaded}
Yilun Chen, Zhicheng Wang, Yuxiang Peng, Zhiqiang Zhang, Gang Yu, and Jian Sun.
\newblock Cascaded pyramid network for multi-person pose estimation.
\newblock In {\em Proceedings of the IEEE conference on computer vision and
  pattern recognition}, pages 7103--7112, 2018.

\bibitem{ci2019optimizing}
Hai Ci, Chunyu Wang, Xiaoxuan Ma, and Yizhou Wang.
\newblock Optimizing network structure for 3d human pose estimation.
\newblock In {\em Proceedings of the IEEE/CVF international conference on
  computer vision}, pages 2262--2271, 2019.

\bibitem{deng2021synchronized}
Yicheng Deng, Cheng Sun, Yongqi Sun, and Jiahui Zhu.
\newblock A synchronized reprojection-based model for 3d human pose estimation.
\newblock {\em arXiv preprint arXiv:2106.04274}, 2021.

\bibitem{drover2018can}
Dylan Drover, Rohith MV, Ching-Hang Chen, Amit Agrawal, Ambrish Tyagi, and Cong
  Phuoc~Huynh.
\newblock Can 3d pose be learned from 2d projections alone?
\newblock In {\em Proceedings of the European Conference on Computer Vision
  (ECCV) Workshops}, pages 0--0, 2018.

\bibitem{fan2022human}
Zhaoxin Fan, Fengxin Li, Hongyan Liu, Jun He, and Xiaoyong Du.
\newblock Human pose driven object effects recommendation.
\newblock {\em arXiv preprint arXiv:2209.08353}, 2022.

\bibitem{fan2022furpe}
Zhaoxin Fan, Yuqing Pan, Hao Xu, Zhenbo Song, Zhicheng Wang, Kejian Wu, Hongyan
  Liu, and Jun He.
\newblock Furpe: Learning full-body reconstruction from part experts.
\newblock {\em arXiv preprint arXiv:2212.00731}, 2022.

\bibitem{fan2022deep}
Zhaoxin Fan, Yazhi Zhu, Yulin He, Qi Sun, Hongyan Liu, and Jun He.
\newblock Deep learning on monocular object pose detection and tracking: A
  comprehensive overview.
\newblock {\em ACM Computing Surveys}, 55(4):1--40, 2022.

\bibitem{gholami2022adaptpose}
Mohsen Gholami, Bastian Wandt, Helge Rhodin, Rabab Ward, and Z~Jane Wang.
\newblock Adaptpose: Cross-dataset adaptation for 3d human pose estimation by
  learnable motion generation.
\newblock In {\em Proceedings of the IEEE/CVF Conference on Computer Vision and
  Pattern Recognition}, pages 13075--13085, 2022.

\bibitem{gong2022posetriplet}
Kehong Gong, Bingbing Li, Jianfeng Zhang, Tao Wang, Jing Huang, Michael~Bi Mi,
  Jiashi Feng, and Xinchao Wang.
\newblock Posetriplet: Co-evolving 3d human pose estimation, imitation, and
  hallucination under self-supervision.
\newblock In {\em Proceedings of the IEEE/CVF Conference on Computer Vision and
  Pattern Recognition}, pages 11017--11027, 2022.

\bibitem{gong2021poseaug}
Kehong Gong, Jianfeng Zhang, and Jiashi Feng.
\newblock Poseaug: A differentiable pose augmentation framework for 3d human
  pose estimation.
\newblock In {\em Proceedings of the IEEE/CVF Conference on Computer Vision and
  Pattern Recognition}, pages 8575--8584, 2021.

\bibitem{goodfellow2020generative}
Ian Goodfellow, Jean Pouget-Abadie, Mehdi Mirza, Bing Xu, David Warde-Farley,
  Sherjil Ozair, Aaron Courville, and Yoshua Bengio.
\newblock Generative adversarial networks.
\newblock {\em Communications of the ACM}, 63(11):139--144, 2020.

\bibitem{guan2021bilevel}
Shanyan Guan, Jingwei Xu, Yunbo Wang, Bingbing Ni, and Xiaokang Yang.
\newblock Bilevel online adaptation for out-of-domain human mesh
  reconstruction.
\newblock In {\em Proceedings of the IEEE/CVF Conference on Computer Vision and
  Pattern Recognition}, pages 10472--10481, 2021.

\bibitem{hao2023divotrack}
Shenghao Hao, Peiyuan Liu, Yibing Zhan, Kaixun Jin, Zuozhu Liu, Mingli Song,
  Jenq-Neng Hwang, and Gaoang Wang.
\newblock Divotrack: A novel dataset and baseline method for cross-view
  multi-object tracking in diverse open scenes.
\newblock {\em arXiv preprint arXiv:2302.07676}, 2023.

\bibitem{hoffmann2019learning}
David~T Hoffmann, Dimitrios Tzionas, Michael~J Black, and Siyu Tang.
\newblock Learning to train with synthetic humans.
\newblock In {\em German conference on pattern recognition}, pages 609--623.
  Springer, 2019.

\bibitem{ionescu2013human3}
Catalin Ionescu, Dragos Papava, Vlad Olaru, and Cristian Sminchisescu.
\newblock Human3. 6m: Large scale datasets and predictive methods for 3d human
  sensing in natural environments.
\newblock {\em IEEE transactions on pattern analysis and machine intelligence},
  36(7):1325--1339, 2013.

\bibitem{jiang2023five}
Jingxia Jiang, Tian Ye, Jinbin Bai, Sixiang Chen, Wenhao Chai, Shi Jun, Yun
  Liu, and Erkang Chen.
\newblock Five a+ network: You only need 9k parameters for underwater image
  enhancement.
\newblock {\em arXiv preprint arXiv:2305.08824}, 2023.

\bibitem{jiang2023back}
Zhongyu Jiang, Zhuoran Zhou, Lei Li, Wenhao Chai, Cheng-Yen Yang, and Jenq-Neng
  Hwang.
\newblock Back to optimization: Diffusion-based zero-shot 3d human pose
  estimation.
\newblock {\em arXiv preprint arXiv:2307.03833}, 2023.

\bibitem{joo2021exemplar}
Hanbyul Joo, Natalia Neverova, and Andrea Vedaldi.
\newblock Exemplar fine-tuning for 3d human model fitting towards in-the-wild
  3d human pose estimation.
\newblock In {\em 2021 International Conference on 3D Vision (3DV)}, pages
  42--52. IEEE, 2021.

\bibitem{kanazawa2018end}
Angjoo Kanazawa, Michael~J Black, David~W Jacobs, and Jitendra Malik.
\newblock End-to-end recovery of human shape and pose.
\newblock In {\em Proceedings of the IEEE conference on computer vision and
  pattern recognition}, pages 7122--7131, 2018.

\bibitem{kocabas2020vibe}
Muhammed Kocabas, Nikos Athanasiou, and Michael~J Black.
\newblock Vibe: Video inference for human body pose and shape estimation.
\newblock In {\em Proceedings of the IEEE/CVF conference on computer vision and
  pattern recognition}, pages 5253--5263, 2020.

\bibitem{kolotouros2019learning}
Nikos Kolotouros, Georgios Pavlakos, Michael~J Black, and Kostas Daniilidis.
\newblock Learning to reconstruct 3d human pose and shape via model-fitting in
  the loop.
\newblock In {\em Proceedings of the IEEE/CVF International Conference on
  Computer Vision}, pages 2252--2261, 2019.

\bibitem{kundu2022uncertainty}
Jogendra~Nath Kundu, Siddharth Seth, Pradyumna YM, Varun Jampani, Anirban
  Chakraborty, and R~Venkatesh Babu.
\newblock Uncertainty-aware adaptation for self-supervised 3d human pose
  estimation.
\newblock In {\em Proceedings of the IEEE/CVF Conference on Computer Vision and
  Pattern Recognition}, pages 20448--20459, 2022.

\bibitem{li2020cascaded}
Shichao Li, Lei Ke, Kevin Pratama, Yu-Wing Tai, Chi-Keung Tang, and Kwang-Ting
  Cheng.
\newblock Cascaded deep monocular 3d human pose estimation with evolutionary
  training data.
\newblock In {\em Proceedings of the IEEE/CVF Conference on Computer Vision and
  Pattern Recognition}, pages 6173--6183, 2020.

\bibitem{lin2021mesh}
Kevin Lin, Lijuan Wang, and Zicheng Liu.
\newblock Mesh graphormer.
\newblock In {\em Proceedings of the IEEE/CVF International Conference on
  Computer Vision}, pages 12939--12948, 2021.

\bibitem{liu2023nighthazeformer}
Yun Liu, Zhongsheng Yan, Sixiang Chen, Tian Ye, Wenqi Ren, and Erkang Chen.
\newblock Nighthazeformer: Single nighttime haze removal using prior query
  transformer.
\newblock {\em arXiv preprint arXiv:2305.09533}, 2023.

\bibitem{liu2022nighttime}
Yun Liu, Zhongsheng Yan, Aimin Wu, Tian Ye, and Yuche Li.
\newblock Nighttime image dehazing based on variational decomposition model.
\newblock In {\em Proceedings of the IEEE/CVF conference on computer vision and
  pattern recognition}, pages 640--649, 2022.

\bibitem{luo2018orinet}
Chenxu Luo, Xiao Chu, and Alan Yuille.
\newblock Orinet: A fully convolutional network for 3d human pose estimation.
\newblock {\em arXiv preprint arXiv:1811.04989}, 2018.

\bibitem{mao2017least}
Xudong Mao, Qing Li, Haoran Xie, Raymond~YK Lau, Zhen Wang, and Stephen
  Paul~Smolley.
\newblock Least squares generative adversarial networks.
\newblock In {\em Proceedings of the IEEE international conference on computer
  vision}, pages 2794--2802, 2017.

\bibitem{martinez2017simple}
Julieta Martinez, Rayat Hossain, Javier Romero, and James~J Little.
\newblock A simple yet effective baseline for 3d human pose estimation.
\newblock In {\em Proceedings of the IEEE international conference on computer
  vision}, pages 2640--2649, 2017.

\bibitem{mehta2017monocular}
Dushyant Mehta, Helge Rhodin, Dan Casas, Pascal Fua, Oleksandr Sotnychenko,
  Weipeng Xu, and Christian Theobalt.
\newblock Monocular 3d human pose estimation in the wild using improved cnn
  supervision.
\newblock In {\em 2017 international conference on 3D vision (3DV)}, pages
  506--516. IEEE, 2017.

\bibitem{mehta2018single}
Dushyant Mehta, Oleksandr Sotnychenko, Franziska Mueller, Weipeng Xu, Srinath
  Sridhar, Gerard Pons-Moll, and Christian Theobalt.
\newblock Single-shot multi-person 3d pose estimation from monocular rgb.
\newblock In {\em 2018 International Conference on 3D Vision (3DV)}, pages
  120--130. IEEE, 2018.

\bibitem{mehta2017vnect}
Dushyant Mehta, Srinath Sridhar, Oleksandr Sotnychenko, Helge Rhodin, Mohammad
  Shafiei, Hans-Peter Seidel, Weipeng Xu, Dan Casas, and Christian Theobalt.
\newblock Vnect: Real-time 3d human pose estimation with a single rgb camera.
\newblock {\em Acm transactions on graphics (tog)}, 36(4):1--14, 2017.

\bibitem{pavllo20193d}
Dario Pavllo, Christoph Feichtenhofer, David Grangier, and Michael Auli.
\newblock 3d human pose estimation in video with temporal convolutions and
  semi-supervised training.
\newblock In {\em Proceedings of the IEEE/CVF Conference on Computer Vision and
  Pattern Recognition}, pages 7753--7762, 2019.

\bibitem{rogez2016mocap}
Gr{\'e}gory Rogez and Cordelia Schmid.
\newblock Mocap-guided data augmentation for 3d pose estimation in the wild.
\newblock {\em Advances in neural information processing systems}, 29, 2016.

\bibitem{shen2018wasserstein}
Jian Shen, Yanru Qu, Weinan Zhang, and Yong Yu.
\newblock Wasserstein distance guided representation learning for domain
  adaptation.
\newblock In {\em Proceedings of the AAAI Conference on Artificial
  Intelligence}, volume~32, 2018.

\bibitem{song2021human}
Liangchen Song, Gang Yu, Junsong Yuan, and Zicheng Liu.
\newblock Human pose estimation and its application to action recognition: A
  survey.
\newblock {\em Journal of Visual Communication and Image Representation},
  76:103055, 2021.

\bibitem{su2017pose}
Chi Su, Jianing Li, Shiliang Zhang, Junliang Xing, Wen Gao, and Qi Tian.
\newblock Pose-driven deep convolutional model for person re-identification.
\newblock In {\em Proceedings of the IEEE international conference on computer
  vision}, pages 3960--3969, 2017.

\bibitem{sun2019deep}
Ke Sun, Bin Xiao, Dong Liu, and Jingdong Wang.
\newblock Deep high-resolution representation learning for human pose
  estimation.
\newblock In {\em Proceedings of the IEEE/CVF conference on computer vision and
  pattern recognition}, pages 5693--5703, 2019.

\bibitem{svenstrup2009pose}
Mikael Svenstrup, Soren Tranberg, Hans~Jorgen Andersen, and Thomas Bak.
\newblock Pose estimation and adaptive robot behaviour for human-robot
  interaction.
\newblock In {\em 2009 IEEE International Conference on Robotics and
  Automation}, pages 3571--3576. IEEE, 2009.

\bibitem{workman2015deepfocal}
Workman \textit{et al.}
\newblock Deepfocal: A method for direct focal length estimation.
\newblock In {\em ICIP}, 2015.

\bibitem{varol2017learning}
Gul Varol, Javier Romero, Xavier Martin, Naureen Mahmood, Michael~J Black, Ivan
  Laptev, and Cordelia Schmid.
\newblock Learning from synthetic humans.
\newblock In {\em Proceedings of the IEEE conference on computer vision and
  pattern recognition}, pages 109--117, 2017.

\bibitem{von2018recovering}
Timo Von~Marcard, Roberto Henschel, Michael~J Black, Bodo Rosenhahn, and Gerard
  Pons-Moll.
\newblock Recovering accurate 3d human pose in the wild using imus and a moving
  camera.
\newblock In {\em Proceedings of the European Conference on Computer Vision
  (ECCV)}, pages 601--617, 2018.

\bibitem{wandt2018kinematic}
Bastian Wandt, Hanno Ackermann, and Bodo Rosenhahn.
\newblock A kinematic chain space for monocular motion capture.
\newblock In {\em Proceedings of the European Conference on Computer Vision
  (ECCV) Workshops}, pages 0--0, 2018.

\bibitem{wandt2019repnet}
Bastian Wandt and Bodo Rosenhahn.
\newblock Repnet: Weakly supervised training of an adversarial reprojection
  network for 3d human pose estimation.
\newblock In {\em Proceedings of the IEEE/CVF Conference on Computer Vision and
  Pattern Recognition}, pages 7782--7791, 2019.

\bibitem{wang2021deep}
Jinbao Wang, Shujie Tan, Xiantong Zhen, Shuo Xu, Feng Zheng, Zhenyu He, and
  Ling Shao.
\newblock Deep 3d human pose estimation: A review.
\newblock {\em Computer Vision and Image Understanding}, 210:103225, 2021.

\bibitem{wang2020predicting}
Zhe Wang, Daeyun Shin, and Charless~C Fowlkes.
\newblock Predicting camera viewpoint improves cross-dataset generalization for
  3d human pose estimation.
\newblock In {\em European Conference on Computer Vision}, pages 523--540.
  Springer, 2020.

\bibitem{wilson2020survey}
Garrett Wilson and Diane~J Cook.
\newblock A survey of unsupervised deep domain adaptation.
\newblock {\em ACM Transactions on Intelligent Systems and Technology (TIST)},
  11(5):1--46, 2020.

\bibitem{yang20183d}
Wei Yang, Wanli Ouyang, Xiaolong Wang, Jimmy Ren, Hongsheng Li, and Xiaogang
  Wang.
\newblock 3d human pose estimation in the wild by adversarial learning.
\newblock In {\em Proceedings of the IEEE conference on computer vision and
  pattern recognition}, pages 5255--5264, 2018.

\bibitem{ye2022towards}
Tian Ye, Sixiang Chen, Yun Liu, Yi Ye, Jinbin Bai, and Erkang Chen.
\newblock Towards real-time high-definition image snow removal: Efficient
  pyramid network with asymmetrical encoder-decoder architecture.
\newblock In {\em Proceedings of the Asian Conference on Computer Vision},
  pages 366--381, 2022.

\bibitem{ye2022underwater}
Tian Ye, Sixiang Chen, Yun Liu, Yi Ye, Erkang Chen, and Yuche Li.
\newblock Underwater light field retention: Neural rendering for underwater
  imaging.
\newblock In {\em Proceedings of the IEEE/CVF Conference on Computer Vision and
  Pattern Recognition}, pages 488--497, 2022.

\bibitem{ye2021perceiving}
Tian Ye, Yunchen Zhang, Mingchao Jiang, Liang Chen, Yun Liu, Sixiang Chen, and
  Erkang Chen.
\newblock Perceiving and modeling density for image dehazing.
\newblock In {\em European Conference on Computer Vision}, pages 130--145.
  Springer, 2022.

\bibitem{zeng2020srnet}
Ailing Zeng, Xiao Sun, Fuyang Huang, Minhao Liu, Qiang Xu, and Stephen Lin.
\newblock Srnet: Improving generalization in 3d human pose estimation with a
  split-and-recombine approach.
\newblock In {\em European Conference on Computer Vision}, pages 507--523.
  Springer, 2020.

\bibitem{zhang2022mixste}
Jinlu Zhang, Zhigang Tu, Jianyu Yang, Yujin Chen, and Junsong Yuan.
\newblock Mixste: Seq2seq mixed spatio-temporal encoder for 3d human pose
  estimation in video.
\newblock In {\em Proceedings of the IEEE/CVF Conference on Computer Vision and
  Pattern Recognition}, pages 13232--13242, 2022.

\bibitem{zhang2023mpm}
Zhenyu Zhang, Wenhao Chai, Zhongyu Jiang, Tian Ye, Mingli Song, Jenq-Neng
  Hwang, and Gaoang Wang.
\newblock Mpm: A unified 2d-3d human pose representation via masked pose
  modeling.
\newblock {\em arXiv preprint arXiv:2306.17201}, 2023.

\bibitem{zhao2019semantic}
Long Zhao, Xi Peng, Yu Tian, Mubbasir Kapadia, and Dimitris~N Metaxas.
\newblock Semantic graph convolutional networks for 3d human pose regression.
\newblock In {\em Proceedings of the IEEE/CVF conference on computer vision and
  pattern recognition}, pages 3425--3435, 2019.

\bibitem{zhao2023survey}
Zhonghan Zhao, Wenhao Chai, Shengyu Hao, Wenhao Hu, Guanhong Wang, Shidong Cao,
  Mingli Song, Jenq-Neng Hwang, and Gaoang Wang.
\newblock A survey of deep learning in sports applications: Perception,
  comprehension, and decision.
\newblock {\em arXiv preprint arXiv:2307.03353}, 2023.

\end{thebibliography}
}

\newsavebox\CBox
\def\textBF#1{\sbox\CBox{#1}\resizebox{\wd\CBox}{\ht\CBox}{\textBF{#1}}}
\newcommand{\ParaEntry}[1]{\vspace{1mm}\noindent\textbf{#1}}
\newcommand{\TableEntry}[2]{#1~\scriptsize{\textcolor{red}{\tiny(-#2)}}}
\newcommand{\BTableEntry}[2]{\textbf{#1}~\scriptsize{\textcolor{red}{\tiny(-#2)}}}
\newcommand{\PTableEntry}[2]{#1~\scriptsize{\textcolor{red}{\tiny(+#2)}}}
\newcommand{\BPTableEntry}[2]{\textbf{#1}~\scriptsize{\textcolor{red}{\tiny(+#2)}}}

\newcommand{\HTableEntry}[2]{#1~\scriptsize{\textcolor{red}{\small(-#2)}}}
\newcommand{\HBTableEntry}[2]{\textbf{#1}~\scriptsize{\textcolor{red}{\small(-#2)}}}
\newcommand{\HPTableEntry}[2]{#1~\scriptsize{\textcolor{red}{\small(+#2)}}}
\newcommand{\HBPTableEntry}[2]{\textbf{#1}~\scriptsize{\textcolor{red}{\small(+#2)}}}

\newpage

\section*{Supplement Material}

\subsection*{A. Experiments of detected 2D pose}
GT 2D pose is used in all the experiments in the paper. Besides, we further evaluate the performance under detected 2D pose as shown in Table~\ref{tab:r33}. Since GPA only uses 2D box rather than specific 2D pose, the performance does not drop a lot. We reimplement all the baseline models and PoseAug in H3.6M.

\renewcommand{\thetable}{A}
\begin{table*}[t]
\centering
\setlength{\tabcolsep}{2mm}
\resizebox{0.9\linewidth}{!}{
\begin{tabular}{l|cccc|cccc}
\toprule
\multicolumn{1}{c|}{} & \multicolumn{4}{c|}{H3.6M}  & \multicolumn{4}{c}{3DHP} \\ 
\midrule
Method & \multicolumn{1}{c}{DET} & \multicolumn{1}{c}{CPN} & \multicolumn{1}{c}{HRNet} & \multicolumn{1}{c|}{GT} &  \multicolumn{1}{c}{DET} & \multicolumn{1}{c}{CPN} & \multicolumn{1}{c}{HRNet} & \multicolumn{1}{c}{GT}\\ 
\midrule
SemGCN & 77.3 & 73.8 & 67.2 & 58.9 & 101.9 & 98.7 & 95.6 & 95.6 \\
+ PoseAug & \HTableEntry{75.5}{1.8}  & \HTableEntry{73.5}{0.3} & \HTableEntry{66.1}{1.2} & \HTableEntry{58.0}{0.9} & \HTableEntry{89.9}{11.9} & \HTableEntry{89.3}{9.4}  & \HTableEntry{89.1}{6.5}  & \HTableEntry{86.1}{9.5} \\
\rowcolor[gray]{0.9}
+ PoseDA & \HBTableEntry{75.0}{2.3} & \HBTableEntry{71.9}{1.9} &  \HBTableEntry{63.8}{3.4}&  \HBTableEntry{53.9}{5.0}& \HBTableEntry{80.3}{21.6} & \HBTableEntry{80.3}{18.4} & \HBTableEntry{80.9}{14.7} & \HBTableEntry{78.3}{17.3}\\
\midrule
SimpleBaseline & 69.2 & 65.1 & 60.3 & 53.6 & 91.1 & 88.8 & 86.4 & 81.2 \\
+ PoseAug & \HTableEntry{68.4}{0.8} & \HTableEntry{64.5}{0.6} & \HTableEntry{59.7}{0.6} & \HTableEntry{51.8}{1.8} & \HTableEntry{78.7}{12.4} & \HTableEntry{78.7}{10.1} & \HTableEntry{76.4}{10.1} & \HTableEntry{76.2}{5.0} \\
\rowcolor[gray]{0.9}
+ PoseDA & \HBTableEntry{67.9}{1.3} & \HBTableEntry{63.0}{2.1}& \HBTableEntry{56.8}{3.5}&\HBTableEntry{50.2}{3.4} &\HBTableEntry{67.3}{23.8} & \HBTableEntry{67.3}{21.5} & \HBTableEntry{67.3}{19.1} & \HBTableEntry{64.7}{16.5}\\
\midrule
ST-GCN & 73.8 & 76.8 & 62.9 & 57.2 & 95.5 & 91.3 & 87.9 & 81.2 \\
+ PoseAug &\HTableEntry{73.8}{0.0} & \HTableEntry{72.9}{3.9} & \HTableEntry{61.3}{1.6} & \HTableEntry{51.2}{6.0} & \HTableEntry{83.5}{12.1} & \HBTableEntry{77.7}{13.6} & \HTableEntry{76.6}{11.3} & \HTableEntry{74.9}{6.3} \\
\rowcolor[gray]{0.9}
+ PoseDA & \HBTableEntry{73.0}{0.8}&\HBTableEntry{68.6}{8.2} &\HBTableEntry{61.2}{1.7} & \HBTableEntry{48.4}{8.8}& \HBTableEntry{78.8}{16.7} & \HTableEntry{77.8}{13.5} & \HBTableEntry{75.1}{12.8} & \HBTableEntry{69.5}{11.7}\\
\midrule
VideoPose3D & 70.4 & 79.2 & 70.7 & 64.7 & 92.6 & 89.8 & 85.6 & 82.3 \\
+ PoseAug & \HBTableEntry{67.1}{3.3} & \HTableEntry{70.4}{8.8} & \HTableEntry{63.6}{7.1} & \HTableEntry{56.7}{8.0} & \HTableEntry{78.3}{14.4} & \HTableEntry{78.4}{11.4} & \HTableEntry{73.2}{12.4} & \HTableEntry{73.0}{9.3}\\
\rowcolor[gray]{0.9}
+ PoseDA & \HTableEntry{67.4}{3.0}&\HBTableEntry{62.2}{17.0} &\HBTableEntry{55.7}{15.0} & \HBTableEntry{49.9}{14.8} & \HBTableEntry{64.6}{28.0} & \HBTableEntry{65.4}{24.4} &  \HBTableEntry{64.5}{21.1} & \HBTableEntry{61.4}{20.9}\\
\bottomrule
\end{tabular}
}
\vspace{10pt}
\caption{{Performance comparison in MPJPE~($\downarrow$) for various pose estimators on H3.6M and 3DHP datasets. DET, CPN, HRNet and GT denote 3D pose estimation model trained on several widely used different 2D pose sources, respectively. For H3.6M Exp., source: H3.6m-S1; target: H3.6m-S5, S6, S7, S8. For 3DHP Exp., source: H3.6M; target: 3DHP.} 
}
\label{tab:r33}
\end{table*}

\subsection*{B. Ablation studies on 3DPW}
We conduct additional ablation studies on 3DPW dataset in Table~\ref{tab:r31}. We believe that the pose diversity is the limitation (\eg the rare poses are still hard to estimate after PoseDA).

\renewcommand{\thetable}{B}
\begin{table}[h]
    \centering
    \small
    \newcommand{\D}[1]{~\scriptsize{\textcolor{red}{(#1)}}}
    \setlength{\tabcolsep}{1mm}
    \resizebox{\linewidth}{!}{
    \begin{tabular}{l|cccc}
        % \specialrule{1pt}{1pt}{2pt}
        \toprule
        Method & MPJPE~($\downarrow$) & PA-MPJPE~($\downarrow$) & PCK~($\uparrow$) & AUC~($\uparrow$)\\
        \midrule
        SimpleBaseline &  153.44 & 100.95 & 59.79 & 28.59\\
        + LPA &  \TableEntry{136.64}{16.8} & \TableEntry{79.07}{21.88} & \PTableEntry{63.07}{3.28} & \PTableEntry{28.99}{0.40}\\
        + GPA &  \TableEntry{131.41}{22.03} & \TableEntry{90.10}{10.85} & \PTableEntry{67.53}{7.74} & \PTableEntry{28.94}{0.35}\\
        \rowcolor[gray]{0.9}
        + PoseDA & \BTableEntry{121.93}{31.51} & \BTableEntry{78.39}{22.56} & \BPTableEntry{69.23}{6.16} & \BPTableEntry{29.72}{0.73}\\
        \midrule
        VideoPose3D & 101.46 & 61.49 & 80.50 & 41.17 \\
        + LPA &  \TableEntry{96.72}{4.74} & \TableEntry{58.96}{2.53} & \PTableEntry{83.42}{2.92} & \PTableEntry{43.17}{2.00}\\
        + GPA &  \TableEntry{92.44}{9.02} & \TableEntry{58.59}{2.90} & \PTableEntry{83.94}{3.44} & \PTableEntry{45.05}{3.88}\\
        \rowcolor[gray]{0.9}
        + PoseDA & \BTableEntry{87.70}{13.76} & \BTableEntry{55.30}{6.19} & \BPTableEntry{84.98}{4.48} & \BPTableEntry{46.10}{4.93}\\
        \bottomrule
    \end{tabular}
    }
    \vspace{10pt}
    \caption{Ablation study on components and pose lifting network of our method. Source: H3.6M. Target: 3DPW.}
    \label{tab:r31}
\end{table}

\subsection*{C. More discussion on Global Position Alignment}

We show that global position alignment module actually align the 2D pose distribution in terms of scale and root position on 2D images in \cref{fig:position}. While other works did camera view estimation~\cite{wandt2019repnet, wang2020predicting} or generation~\cite{gholami2022adaptpose, gong2021poseaug} as an auxiliary task to address the global position adaptation problems, our method achieves alignment explicitly and directly.

\renewcommand{\thefigure}{A}
\begin{figure}[h]
    \centering
    \includegraphics[width=0.95\linewidth]{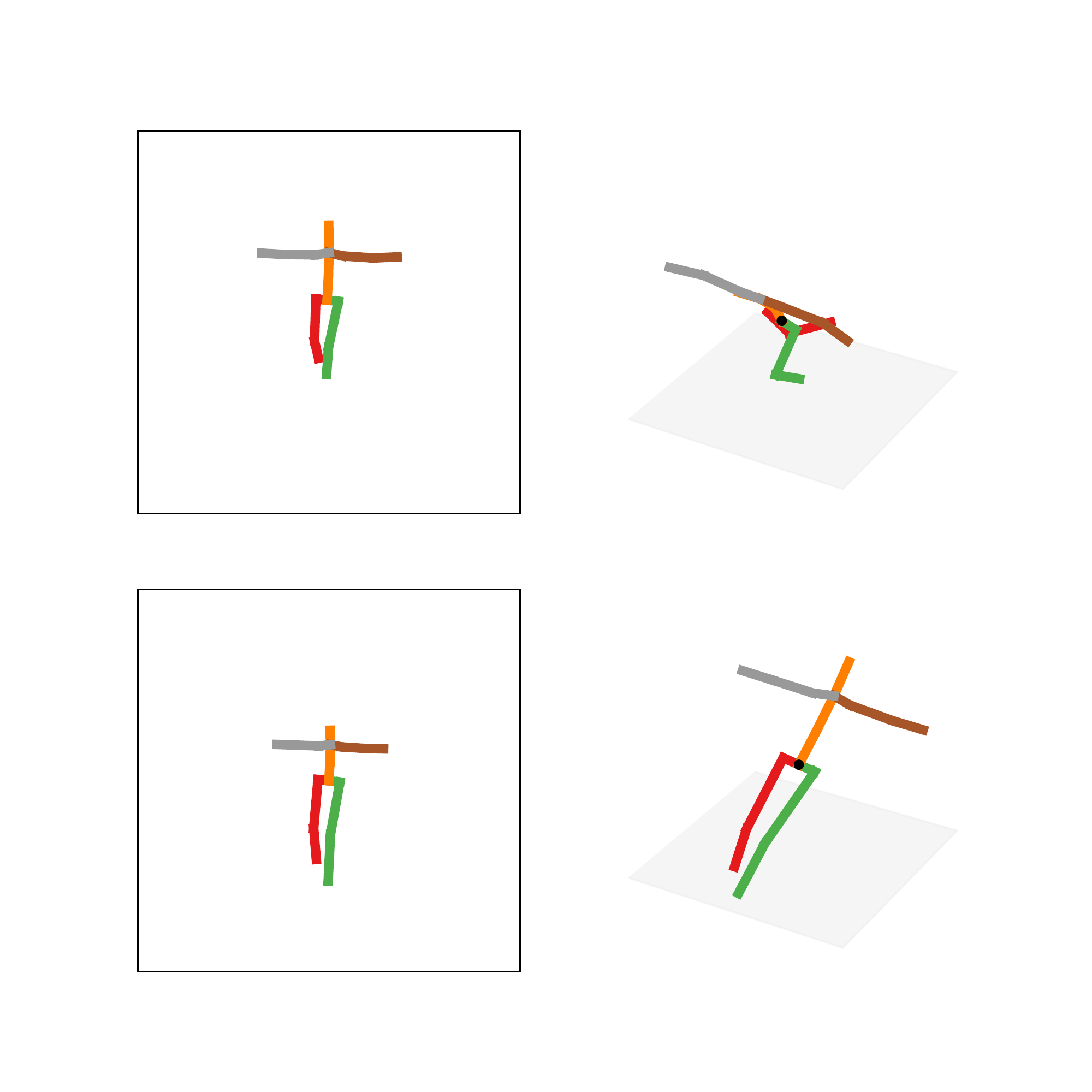}
    \caption{These 2 sampled poses are from generated poses and the target dataset. They have similar 2D poses but different 3D poses, indicating that adaptation based on 2D poses may not lead to adaptation on 3D poses.}
    \label{fig:2d3dcomparison}
\end{figure}

\renewcommand{\thefigure}{B}
\begin{figure*}[t]
    \centering
    \includegraphics[width=0.95\linewidth]{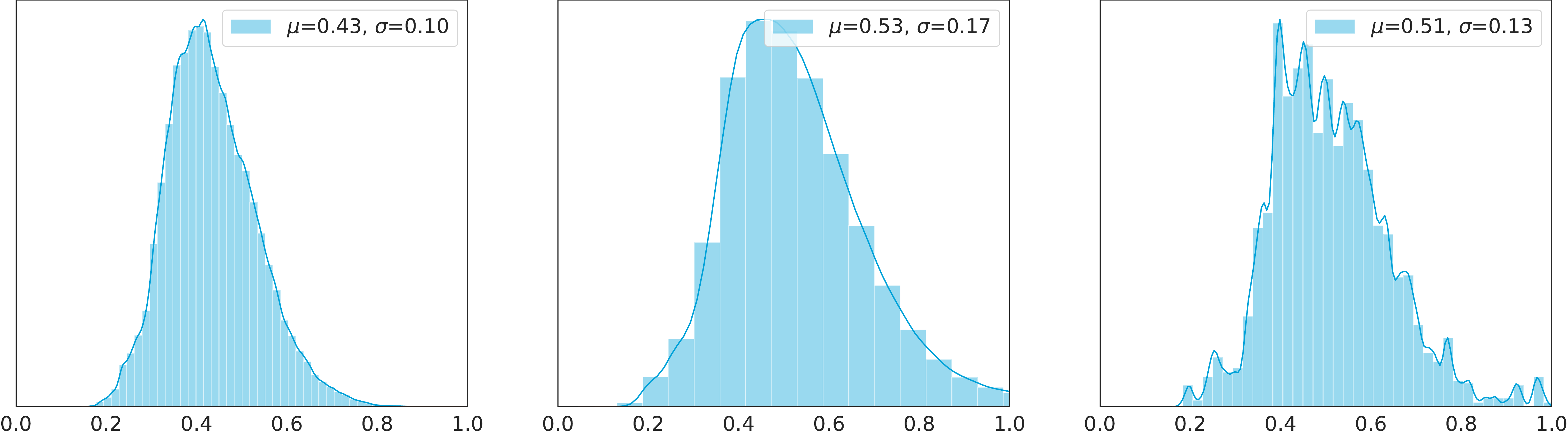}
    \includegraphics[width=0.95\linewidth]{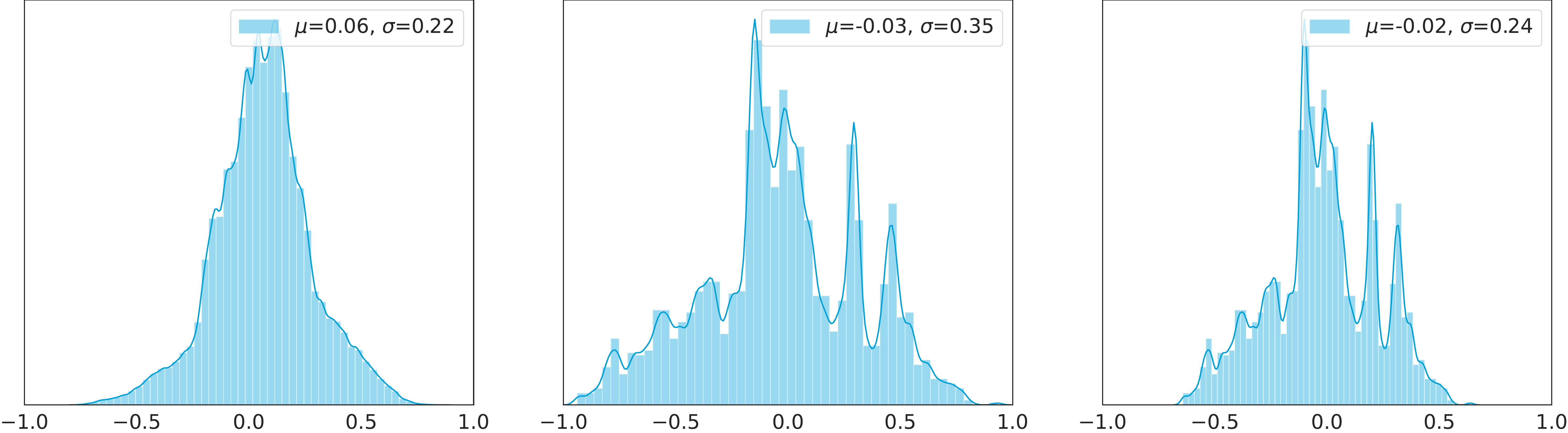}
    \includegraphics[width=0.95\linewidth]{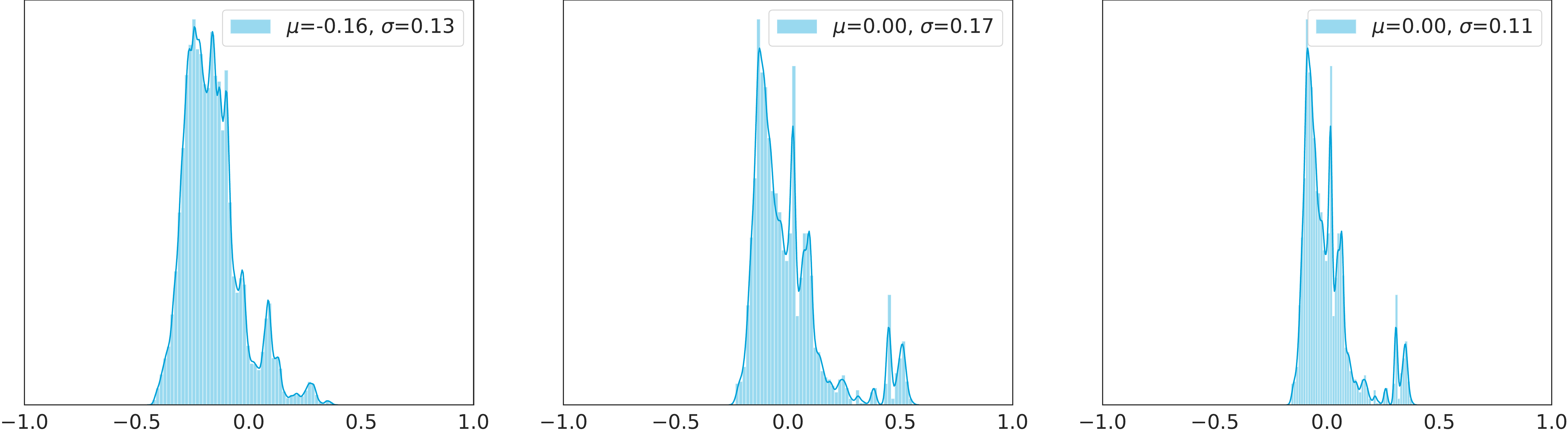}
    \caption{Comparison of 2D scale (first row), root position of x-axis (second row), y-axis (third row) in source domain (left), source domain after GPA (middle), target domain (right). Source: H3.6M. Target: 3DHP.}
    \label{fig:position}
\end{figure*}

\subsection*{D. More discussion on Local Pose Augmentation}

The most counter-intuitive conclusion in this paper is why adaptation methods perform worse than augmentation methods. In the discussion section, we include more detailed ablation studies on LPA. As shown in \cref{fig:2d3dcomparison}, we sampled two poses from generated poses trained with a 2D discriminator and the target dataset. They have similar 2D poses but different 3D poses, which shows the reason why simply applying local pose adaptation based on a 2D pose discriminator may not have the final adaptation performance.

\renewcommand{\thetable}{C}
\begin{table}[h]
    \centering
    \setlength{\tabcolsep}{1mm}
    
    \begin{tabular}{ccc|ccc}
        \toprule
        $\mathcal{G}_{pose}$ & $\mathcal{D}_{3D}$ & $\mathcal{D}_{2D}$ & MPJPE~($\downarrow$) & PCK~($\uparrow$) & AUC~($\uparrow$)\\
        \midrule
        - & - & - & 66.07 & 90.87 & 60.07\\
        $\mathcal{S}$ & $\mathcal{S}$ & $\mathcal{S}$ & 73.55 & 88.96 & 56.41\\
        $\mathcal{S}$ & $\mathcal{S}$ & $\mathcal{T}$ & 65.46 & 91.27 & 60.03\\
        $\mathcal{S}$ & $\mathcal{S}$ & - & \textbf{61.36}& \textbf{92.05} & \textbf{62.52}\\
        \bottomrule
    \end{tabular}
    \vspace{10pt}
    \caption{The input of the pose generator $\mathcal{G}_{pose}$, the 3D pose discriminator $\mathcal{D}_{3D}$, and the 2D pose discriminator $\mathcal{D}_{2D}$ in Local Pose Augmentation (LPA) module. $S, T$ denote poses from the source or target domain. Source: H3.6M. Target: 3DHP.}
    \label{tab:dis}
\end{table}

As \cref{tab:dis}, compared with our final method, the 2D discriminator trained with 2D poses from the target dataset improves the performance from $66.07$ mm to $65.46$ mm in MPJPE since the discriminator makes scale and location adaptation better. However, once the 2D discriminator is removed, we can achieve a better result, $61.36$ mm. The reason is that the 2D pose discriminator suppresses the diversity of generated 3D poses and makes the generator generates poses with similar 2D poses, but different 3D poses, as \cref{fig:2d3dcomparison} shows.

\end{document}